\documentclass[11pt,a4paper]{article}
\usepackage[hyperref]{emnlp-ijcnlp-2019}

\usepackage{times}
\usepackage{latexsym}

\usepackage{todonotes}
\usepackage{subcaption}

\usepackage{caption}
\usepackage{amsmath,amssymb}
\usepackage{booktabs, longtable, supertabular}
\usepackage{graphicx}
\usepackage{multirow}

\usepackage{tikz,graphics,color,float,epsf}

\newcommand{\R}{\mathbb{R}}

\usepackage{url}

\aclfinalcopy 

\title{Investigating Multilingual NMT Representations at Scale}

\author{
Sneha Kudugunta \thanks{ \hspace{0.1cm} Work done as part of Google AI Residency} \qquad
  Ankur Bapna \qquad
  Isaac Caswell \AND
  \vspace{0.5cm}
  Naveen Arivazhagan\qquad 
  Orhan Firat \\
  {\centering Google AI} \\ 
  {\centering \tt {\{snehark,ankurbpn,icaswell,navari,orhanf\}}@google.com}
  }

\date{}

\begin{document}
\maketitle

\begin{abstract}

Multilingual Neural Machine Translation (NMT) models have yielded large empirical success in transfer learning settings. However, these black-box representations are poorly understood, and their mode of transfer remains elusive. In this work, we attempt to understand massively multilingual NMT representations (with 103 languages) using Singular Value Canonical Correlation Analysis (SVCCA), a representation similarity framework that allows us to compare representations across different languages, layers and models. Our analysis validates several empirical results and long-standing intuitions, and unveils new observations regarding how representations evolve in a multilingual translation model. We draw three major conclusions from our analysis, with implications on cross-lingual transfer learning: (i) Encoder representations of different languages cluster based on linguistic similarity, (ii) Representations of a source language learned by the encoder are dependent on the target language, and vice-versa, and (iii) Representations of high resource and/or linguistically similar languages are more robust when fine-tuning on an arbitrary language pair, which is critical to determining how much cross-lingual transfer can be expected in a zero or few-shot setting. We further connect our findings with existing empirical observations in multilingual NMT and transfer learning.

\end{abstract}

\section{Introduction}\label{sect:intro}
Multilingual Neural Machine Translation (NMT) models have demonstrated great improvements for cross-lingual transfer, on tasks including low-resource language translation \cite{zoph2016transfer,nguyen-chiang:2017:I17-2,neubig2018rapid} and zero or few-shot transfer learning for downstream tasks \cite{eriguchi2018zero,lample2019cross,DBLP:journals/corr/abs-1904-09077}. A possible explanation is the ability of multilingual models to encode text from different languages in a shared representation space, resulting in similar sentences being aligned together \cite{firat2016multi,johnson2017google,aharoni2019massively,arivazhagan2019massively}. This is justified by the success of multilingual representations on tasks like sentence alignment across languages \cite{artetxe2018massively}, zero-shot cross-lingual classification \cite{eriguchi2018zero} and XNLI \cite{lample2019cross}.
Although there exist empirical results that suggest that transfer is stronger across similar languages \cite{zoph2016transfer,neubig2018rapid,DBLP:journals/corr/abs-1904-09077}, the mechanism of generalization in multilingual representations is poorly understood.

While interpretability is still a nascent field, there has been some work on investigating the learning dynamics and nature of representations learnt by neural models \cite{olah2018building}. 
Singular Value Canonical Correlation Analysis (SVCCA) is one such method that allows us to analyze the similarity between noisy, high-dimensional representations of the same data-points learnt across different models, layers and tasks \cite{raghu2017svcca}. SVCCA has been used to understand the learning dynamics and representational similarity in a variety of computer vision \cite{Morcos:2018:IRS:3327345.3327475}, language models \cite{saphra2018understanding} and NMT \cite{bau2018identifying}.

In this work, we attempt to peek into the black-box of massively multilingual NMT models, trained on 103 languages, with the lens of SVCCA. We attempt to answer:
\begin{itemize}
\item Which factors determine the extent of overlap in the learned representations?
\item Is the extent of representational overlap similar throughout the model?
\item How robust are multilingual NMT representations to fine-tuning on an arbitrary other language?
\end{itemize}

Answers to the above questions might have large implications on how we approach multilingual models and cross lingual transfer learning. Our work is the first that attempts to understand the nature of multilingual representations and cross-lingual transfer in deep neural networks, based on analyzing a model trained on 103 languages simultaneously.

We structure the study into these sections: In Section \ref{sect:setup}, we describe the experimental setup and tools used to train and analyze our multilingual NMT model. Following that, we attempt to answer each of the above questions in Sections \ref{sec:langsim} and \ref{sect:ft}. Finally in Section~\ref{sect:impl} we summarize our findings with a discussion of the potential implications.\footnote{Tools for online visualization and representation similarity used in our work will be open-sourced to facilitate further analysis.}

\section{Experimental Setup}\label{sect:setup}

\subsection{Data and Model}\label{sub:data}

We  study  multilingual NMT on a massive scale, using an in-house training corpus \cite{arivazhagan2019massively} generated by crawling and extracting parallel sentences from the web \cite{uszkoreit2010large}. Our dataset contains more than 25 billion sentence pairs for 102 languages to and from English, adding up to 204 direct language pairs. 

Having being crawled from the web, our dataset has some important characteristics worth mentioning.

\begin{figure}[t!]
\begin{center}
\includegraphics[scale=0.35]{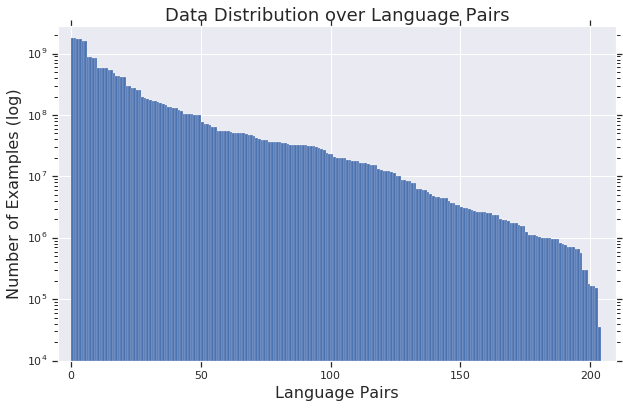}
\caption{Per language pair data distribution of the dataset used to train our multilingual model. The y-axis depicts the number of training examples available per language pair on a logarithmic scale. Dataset sizes range from $10^4$ for the lowest resource language pairs to $10^9$  for the largest.}
\label{fig:data}
\end{center}
\end{figure}

\begin{enumerate}
    \item \textbf{Heavy imbalance between language pairs:} The number of parallel sentences per language pair ranges between $10^4$ to $10^9$. Figure \ref{fig:data} illustrates the data distribution for all the language pairs we study. Although this skew introduces optimization challenges (see Appendix~\ref{sec:model_details}), it also creates a plausible setup for maximizing the positive language transfer from high-resource to low-resource language pairs, making it possible to study low-resource languages, that would otherwise have been very low quality.\footnote{We provide baseline BLEU scores for all languages in Appendix~\ref{sub:bleu_orig}, notice the improvements for low-resource languages in our multilingual setup.} 
    \item \textbf{Diversity:} Our corpus has languages belonging to a wide variety of scripts and linguistic families. These characteristics of our dataset make the problem that we study as realistic as possible. Models trained on this massive open-domain dataset are expected to yield rich, complex representations which we attempt to study in this paper. 
\end{enumerate}

To minimize confounding factors and control the evaluation set size and domain, we created a multi-way aligned  evaluation set containing nearly 3k sentence pairs for all languages.\footnote{Each sentence in our evaluation set is semantically identical across all other languages.} This also allows us to analyze the representational difference and similarity while controlling for semantics.

We use the Transformer-Big \cite{vaswani2017attention} architecture containing 375M parameters described in \citep{chen-EtAl:2018:Long1,arivazhagan2019massively} for our experiments and share all parameters across language pairs including softmax layer and input/output word embeddings. For vocabulary, we use a Sentence Piece Model \cite{kudo2018sentencepiece} with 64k tokens shared on both the encoder and decoder side. 

Each set of parallel sentences has a \verb <2xx>  token  prepended to the source sentence to indicate the target language, as in \cite{johnson2017google}. \footnote{Further details of the model and training routines are given in Appendix~\ref{sec:model_details}.}

\subsection{SVCCA}\label{sub:svcca}

In this study we use Singular Value Canonical Correlation Analysis (SVCCA) \cite{raghu2017svcca} as our investigation tool. SVCCA is a technique to compare vector representations in a way that is both invariant to affine transformations and fast to compute. We express a layer's representations as its activations on a set of $n$ examples, $X = \{x_1, \ldots, x_n\}$. Let $l_1 \in \R^{n \times d_1}$ and  $l_2 \in \R^{n \times d_2}$ be the representations of two neural network layers, with $d_1$ and $d_2$ being the dimensions of the layers corresponding to $l_1$ and $l_2$ respectively. Given layer activations over the set $X$, SVCCA does the following:

\begin{enumerate}
    \item Computes SVD decompositions of $l_1$ and $l_2$ to get subspaces $l'_1  \in \R^{n \times d'_1}$ and $l'_2  \in \R^{n \times d'_2}$ that capture most of the variance in the subspace.\footnote{We retain $99\%$ of the variance in our studies.} 
    \item Uses Canonical Correlation Analysis (CCA) \cite{hardoon2004canonical} to linearly transform $l'_1$ and $l'_2$ to be as aligned as possible, i.e., CCA computes $\tilde{l_1} = W_1 l'_1$ and $\tilde{l_2} = W_2 l'_2$ to maximize the correlations $\bar{\rho} = \{ \rho_1, ..., \rho_{min(d'_1, d'_2)} \}$ between the new subspaces.
\end{enumerate}

As done in \cite{raghu2017svcca}, we use the mean of the correlations, $\bar{\rho}$, as an approximate measure of the relatedness of representations.

\subsubsection*{SVCCA for Sequences}

Recent work on interpretability for NLU tasks uses methods such as diagnostic tasks \cite{belinkov2017neural, tenney2019you,belinkov2018evaluating}, attention based methods \cite{raganato2018analysis} or  task analysis \cite{zhang2018language} and is primarily focused on understanding the linguistic features encoded by a trained model. Some recent works compare learned language vectors \cite{ostling2016continuous, tan2019multilingual,tiedemann2018emerging} and find conclusions similar to ours. To the best of our knowledge, we are the first to compare the hidden representations of the sentences themselves.

Some recent work has applied SVCCA (or CCA) to language modeling \cite{Morcos:2018:IRS:3327345.3327475,saphra2018understanding, dalvi2019one} and NMT \cite{bau2018identifying, dalvi2019one}. However, while \citet{raghu2017svcca} compare the learning dynamics of different classes in an image classifier, to the best of our knowledge, we are the first to apply SVCCA to a multilingual NMT or multitask setting. SVCCA was originally proposed for feed-forward neural networks, but our domain requires comparing unaligned sequences in different languages.

\citet{sahbi2018learning} proposes an alignment agnostic CCA approach to comparing unaligned data. However, the likelihood matrix $D$ specifying alignment of datapoints across different datasets (say, language $A$ and $B$) is application specific and infeasible to obtain in a multilingual setting. A simple heuristic is to summarize a set of activations by applying a pooling operation over the set. This is equivalent to assuming that a given token in a sentence from language $A$ is equally likely to be aligned to each token in an equivalent sentence in language $B$. We perform SVCCA on the hidden representations of the model, averaged over sequence time-steps, for each sentence in our evaluation set. We compare this approach with a naive token level SVCCA strategy in \ref{sub:svcca-seq}. 

\subsubsection*{SVCCA Across Languages}
In all known work using SVCCA, representations of the same data are used for analysis. However, in order to compare representations across languages, we leverage our multi-way parallel evaluation set to compare representations across different languages, since each data point is semantically equivalent across languages.

\begin{figure*}[t!]
  \centering
  \includegraphics[width=.85\textwidth]{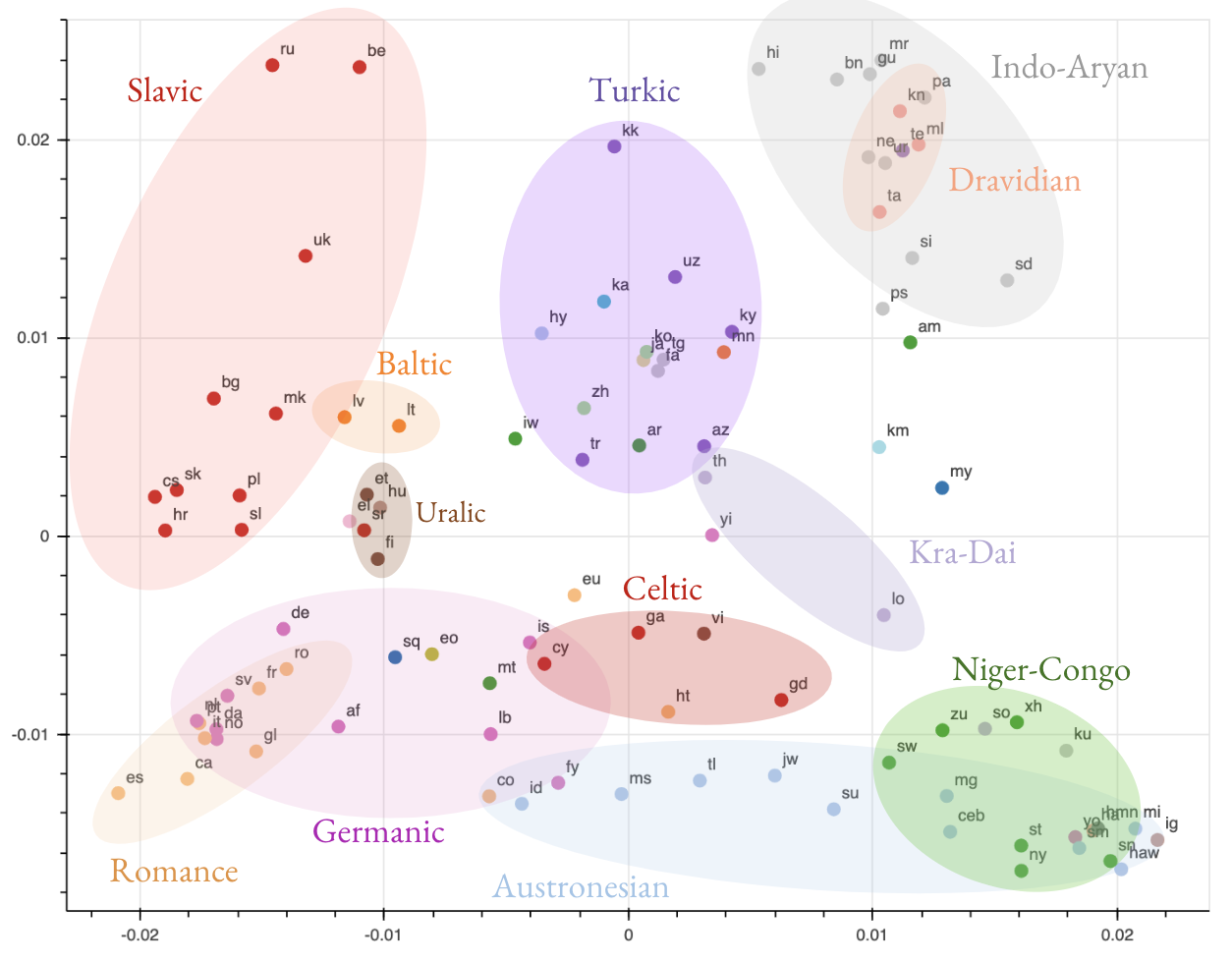}
  \caption{Visualizing clustering of the encoder representations of all languages, based on their SVCCA similarity. Languages are color-coded by their linguistic family. Best viewed in color.}
  \label{fig:allcluster}
\end{figure*}

\section{Multilingual NMT Learns Language Similarity}\label{sec:langsim}

In this section, we use SVCCA to examine the relationship between representations of different languages learned by our massively multilingual NMT model. We compute SVCCA scores of layer-wise activations of a fully trained model between 103 languages pairs in both the Any-to-English and English-to-Any directions.\footnote{Our multilingual NMT model is trained on the available training data which is English centric, hence an All-to-All multilingual model internally decomposes into All-to-English (X-En) and English-to-All (En-X) translation bundles, excluding zero-shot directions.}

\begin{figure}[t!]
\begin{center}
\includegraphics[scale=0.13]{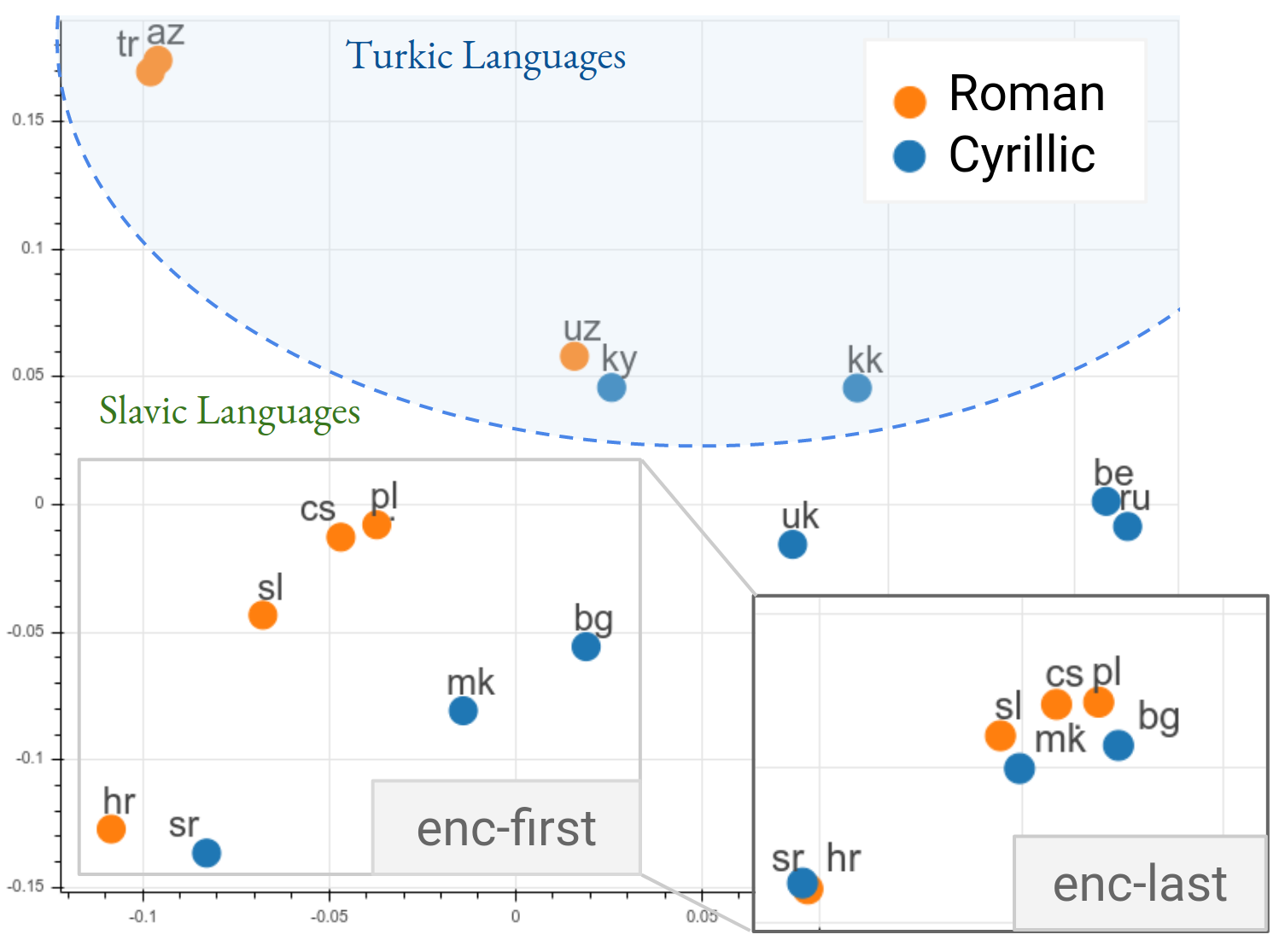}
\caption{Representations of the Turkic and Slavic languages at the subword embeddings, compared with the top of the encoder, and colored by script. The inset shows a portion of the same image at the top of the encoder, focusing on the South-Western Slavic languages. Both images are at the same scale, see Appendix-Fig.~\ref{fig:slavoturkic-clusters} for more details. Best seen in color.}
\label{fig:script}
\end{center}
\end{figure}

\paragraph{Visualizing the Representations}\label{sub:vis}
We first visualize the relationship between languages in their representation space for each layer using Spectral Embeddings \cite{belkin2003laplacian} \footnote{We use the Laplacian Eigenmaps implementation of \href{https://scikit-learn.org/stable/modules/generated/sklearn.manifold.SpectralEmbedding.html}{Spectral Embeddings} in scikit-learn as of August 2019.}. In our case, we use mean SVCCA scores described in Section \ref{sub:svcca} as a similarity measure. Due to the differing nature of translating multiple languages to English and vice versa, the representation space of these two sets of languages, All-to-English and English-to-Any, behave differently and separate quite clearly (Figure \ref{fig:xx-clusters} in the Appendix). We first visualize the encoder representation of all languages in the All-to-English language pair set in Figure \ref{fig:allcluster}. For the sake of analysis, we then visualize subsets of the aforementioned 103 languages in Figures \ref{fig:tree} and \ref{fig:script}. We include visualizations of representations extracted from the embeddings and top layers of the encoder and decoder in the Appendix.

\subsection{What is Language Similarity?} \label{sub:language-similarity}
\begin{figure*}[t!]
\begin{center}
\includegraphics[scale=0.15]{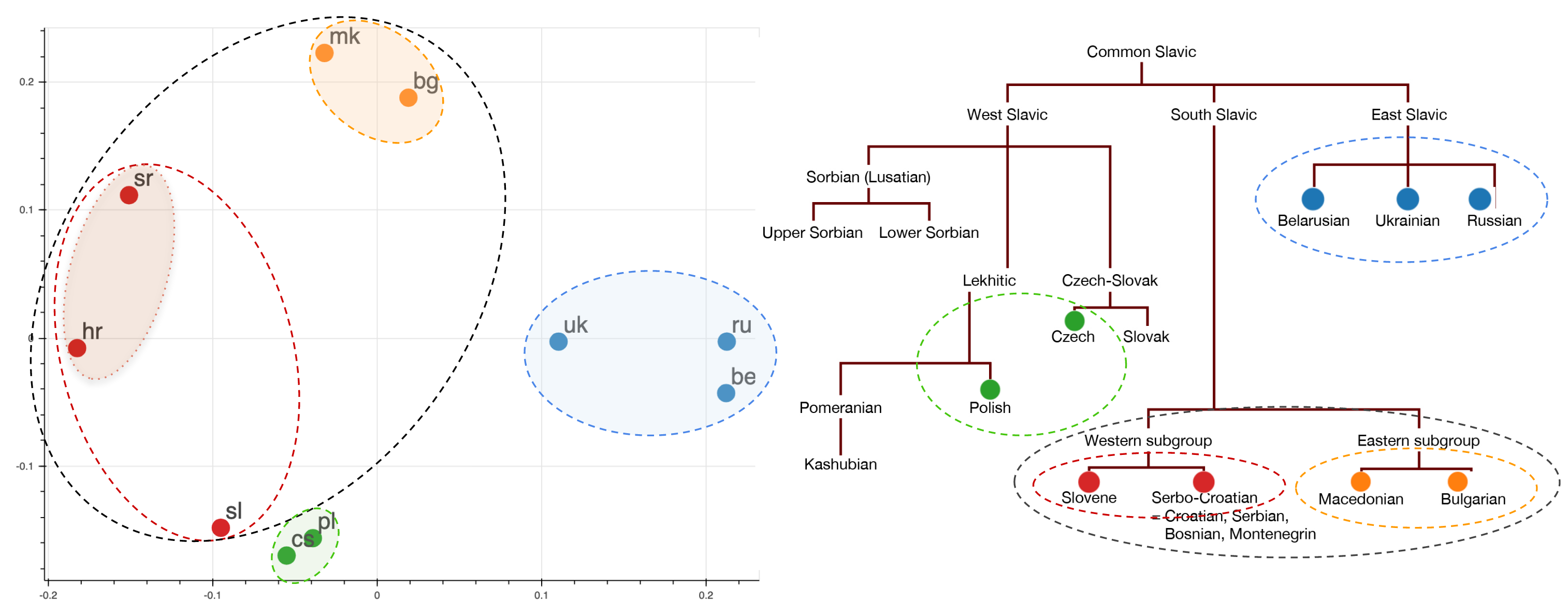}
\caption{Comparing clusterings in the subword embeddings of the Slavic languages in our dataset with their family tree, which is the result of centuries of scholarship by linguists (\cite{slavtree}). Best seen in color.}
\label{fig:tree}
\end{center}
\end{figure*}

In the following sections we draw comparisons between the representational similarity of languages learned by our models, and the linguistic similarity between those languages.
While there are entire sub-fields in linguistics devoted to studying similarity (e.g. Comparative Linguistics and Linguistic Typology \cite{brittanicalanguageclassification}), in this paper we define language similarity in terms of membership in the same language family (e.g. Turkic languages), or branch within that family (e.g. Oghuz Turkic languages). Families are groups of languages believed to share a common ancestor, and therefore tend to have similar vocabulary and grammatical constructs. An example phylogenetic language tree is given in Figure \ref{fig:tree}, on the right.




We also discuss writing systems, including scripts like Cyrillic, Roman, and Ge'ez. While similar languages frequently share the same script, that is not always true. Note that all of these categories\footnote{When we refer to languages from a certain category, we only list those that are in our dataset. For example, when listing Turkic languages we exclude Bashkir, because we do not have English-Bashkir parallel data.} are muddled by various factors that are difficult to tease apart, and might be affected by the web-crawled data that we train on. For instance, languages sharing a script may also be part of the same political bloc, influencing what text is on the web. This and other confounding factors make a rigorous comparison exceedingly difficult. For brevity, we label languages in images with their BCP-47 language codes \cite{bcp47}, which are enumerated in the Appendix, Table \ref{tab:langids}.



\subsection{Representations cluster by language similarity} \label{section:linguistic-clusters}
We first visualize a clustering for all languages together in Figure \ref{fig:allcluster}. While there are a few outliers, we can observe some overlapping clusters, including the Slavic cluster on the top-left, the Germanic and Romance clusters on the bottom-left, the Indo-Aryan and Dravidian clusters on the top-right, etc. To analyze language clustering in more detail we visualize sub-sets of the above languages.

In Figures \ref{fig:tree} and \ref{fig:script}, we visualize the Slavic and Turkic languages in our dataset.
These languages come from two distinct families with very different linguistic properties, and within each family there are languages that are written in Cyrillic and Roman alphabets. This makes them ideal for understanding the interaction between superficial similarity (having the same alphabet and thus sharing many subwords) and linguistic similarity (sharing similar grammatical properties).

The first remarkable phenomenon we observe is that the clusters resulting from our model are grouped not only by family (Slavic), but branches within it (e.g. South Slavic), branches within those branches (e.g. Western Subgroup), and dialects within those (e.g. Serbo-Croatian). Figure \ref{fig:tree} provides a more detailed look into the Slavic languages, and how this compositionality maps to the established family tree for Slavic languages. As can be seen in Figure \ref{fig:script}, this phenomenon can also be observed for Turkic languages, with the Oghuz languages (Turkish and Azeri) forming one cluster, and the two Eastern branches Kipchak and Karluk (Uzbek, Kyrgyz, Kazakh) forming another.

\begin{figure*}[ht!]
\begin{center}
\begin{minipage}{0.5\textwidth}
\includegraphics[width=1.0\textwidth]{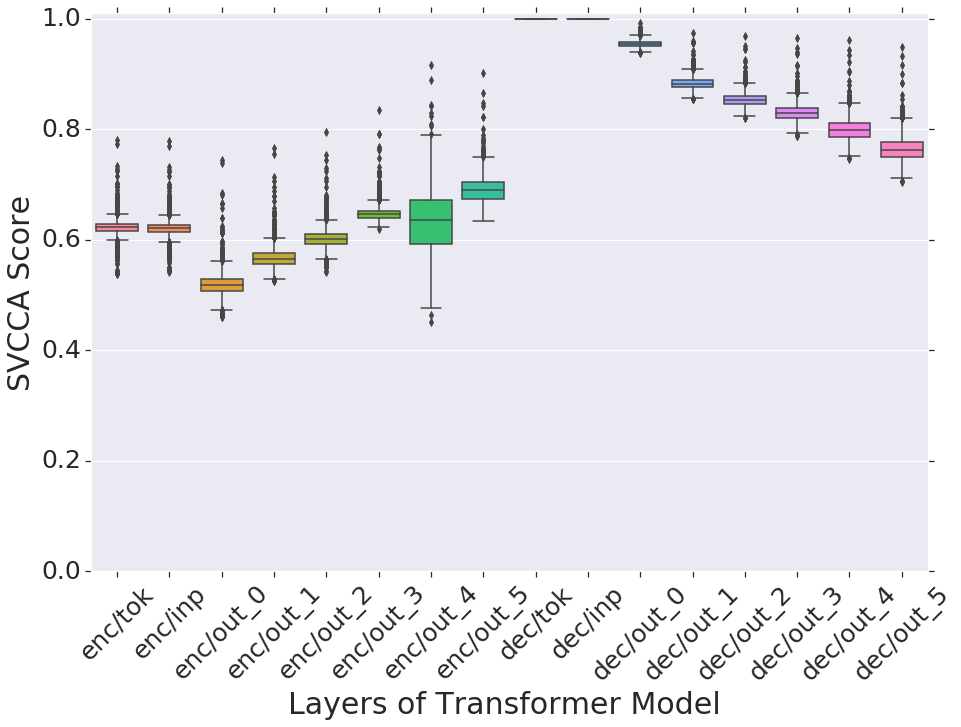}
\subcaption[]{X-En Language Pairs}\label{fig:xen-var}
\end{minipage}%
\begin{minipage}{0.5\textwidth}
\includegraphics[width=1.0\textwidth]{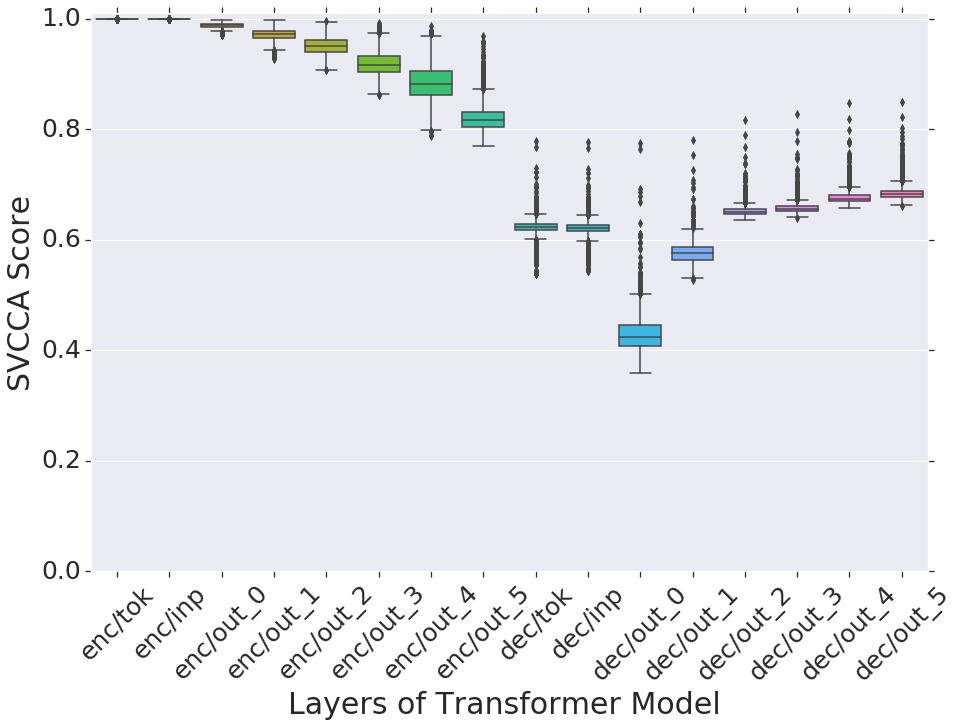}
\subcaption[]{En-X Language Pairs}\label{fig:enx-var}
\end{minipage}%
\end{center}
\caption{The change in distribution of pairwise SVCCA scores between language pairs across layers of a multilingual NMT model, with SVCCA scores between English-to-Any and Any-to-English language pairs visualized separately. We see that while the encoder in (a) and decoder in (b) have dissimilar representations across languages,  the English representations of the decoder in (a) and the encoder in (b) diverge depending on the language X.}
\end{figure*}\label{fig:variance}

A point worth special notice is the closeness between Serbian (sr) and Croatian (hr). Although these two are widely considered registers of the same language \cite{slaviclangs}, Serbian is written in Cyrillic, whereas Croatian is written in the Roman script. However, we see in both Figure \ref{fig:tree} (middle left of plot) and Figure \ref{fig:script} (bottom left of plot) that they are each others' closest neighbors. Since they have no overlap in subword vocabulary, we conclude that they cluster purely based on distributional similarity -- even at the level of sub-word embeddings.

Although we see strong clustering by linguistic family, we also notice the importance of script and lexical overlap, especially (and unsurprisingly) in the embeddings. In Figure \ref{fig:script} we visualize the Turkic and Slavic languages, and color by script. Although the linguistic cluster is probably stronger, there is also a distinct grouping by script, with the Roman-scripted languages on the left and the Cyrillic-scripted languages on the right. However, as we move up the encoder, the script associations become weaker and the language family associations become stronger. The inset in Figure \ref{fig:script} shows the seven South-Western Slavic languages at the top of the encoder, where they have clustered closer together. Again, Serbian and Croatian are an excellent example: \textit{by the top of the encoder, they have become superimposed, diminishing the effect of the difference in script}.

Our conclusions are similar to that of works that have attempted to cluster learned language vectors: \citet{ostling2016continuous, tan2019multilingual} both find that hierarchical clusters of language vectors discover linguistic similarity, with the former finding fine-grained clusterings for Germanic languages. In a similar vein,  \citet{tiedemann2018emerging} visualizes language vectors and find that they roughly cluster by linguistic family.

We find that the trends discussed above are generally true for other language groupings too. The Appendix shows an example with the Dravidian, Indo-Aryan, and Iranian language families, demonstrating the same phenomena discussed above (Appendix Figure~\ref{fig:indo-iranian-dravidian-clusters}). Section \ref{sub:nn} further analyzes how the nearest neighbors of languages in SVCCA space become more linguistically coherent as one moves up the encoder.

\begin{figure*}[t!]
\begin{subfigure}{1.0\textwidth}
  \centering
  \includegraphics[width=1.0\textwidth]{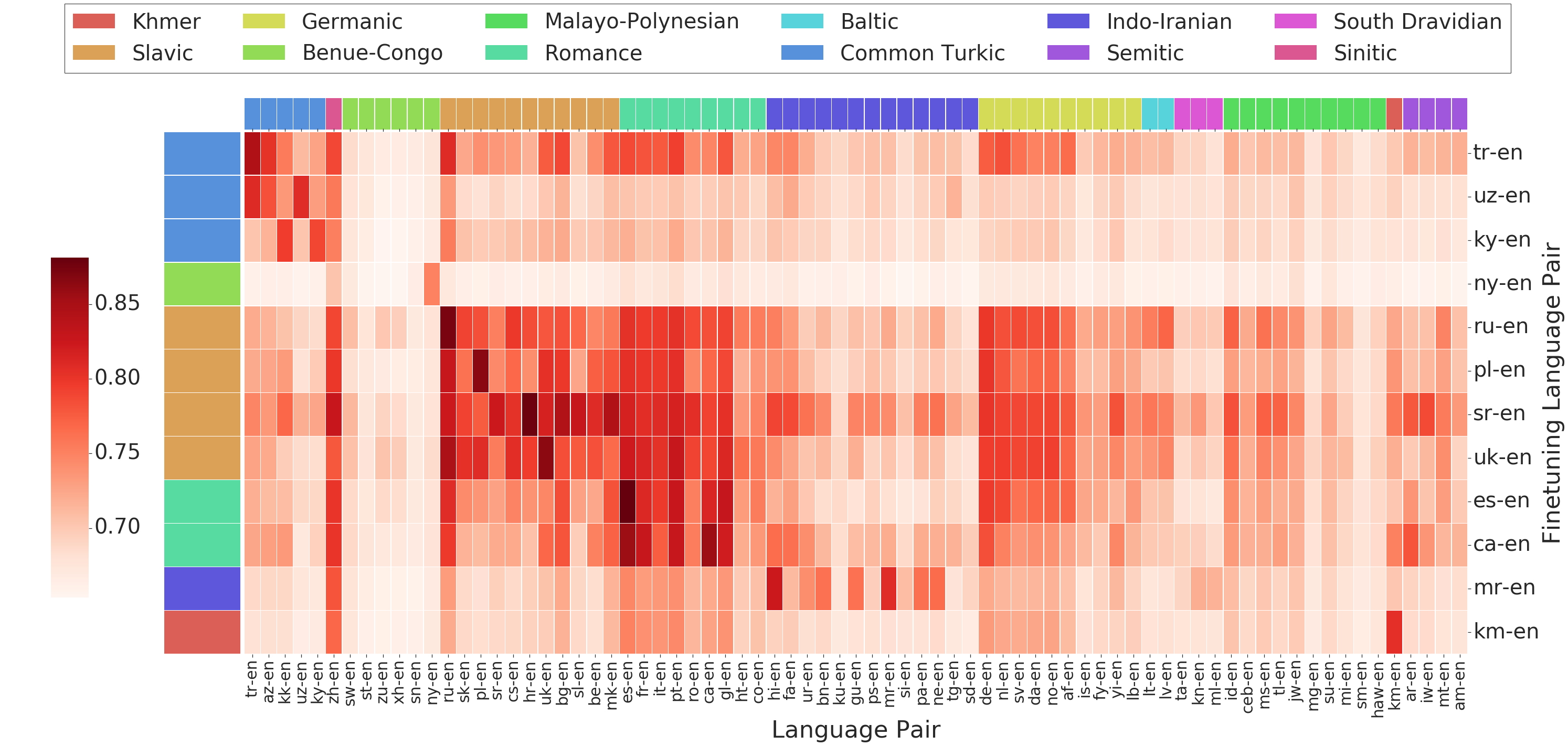}
\caption{SVCCA scores between the representations (top encoder layer) of  \textit{xx-en}  language pairs before and after finetuning on various X-En language pairs.  Darker cell means less change in representation (and higher SVCCA score) of \textit{xx-en} on finetuning with X-En.}
\label{fig:ft-xen}
\end{subfigure}

\begin{subfigure}{1.0\textwidth}
\centering
\includegraphics[width=1.0\textwidth]{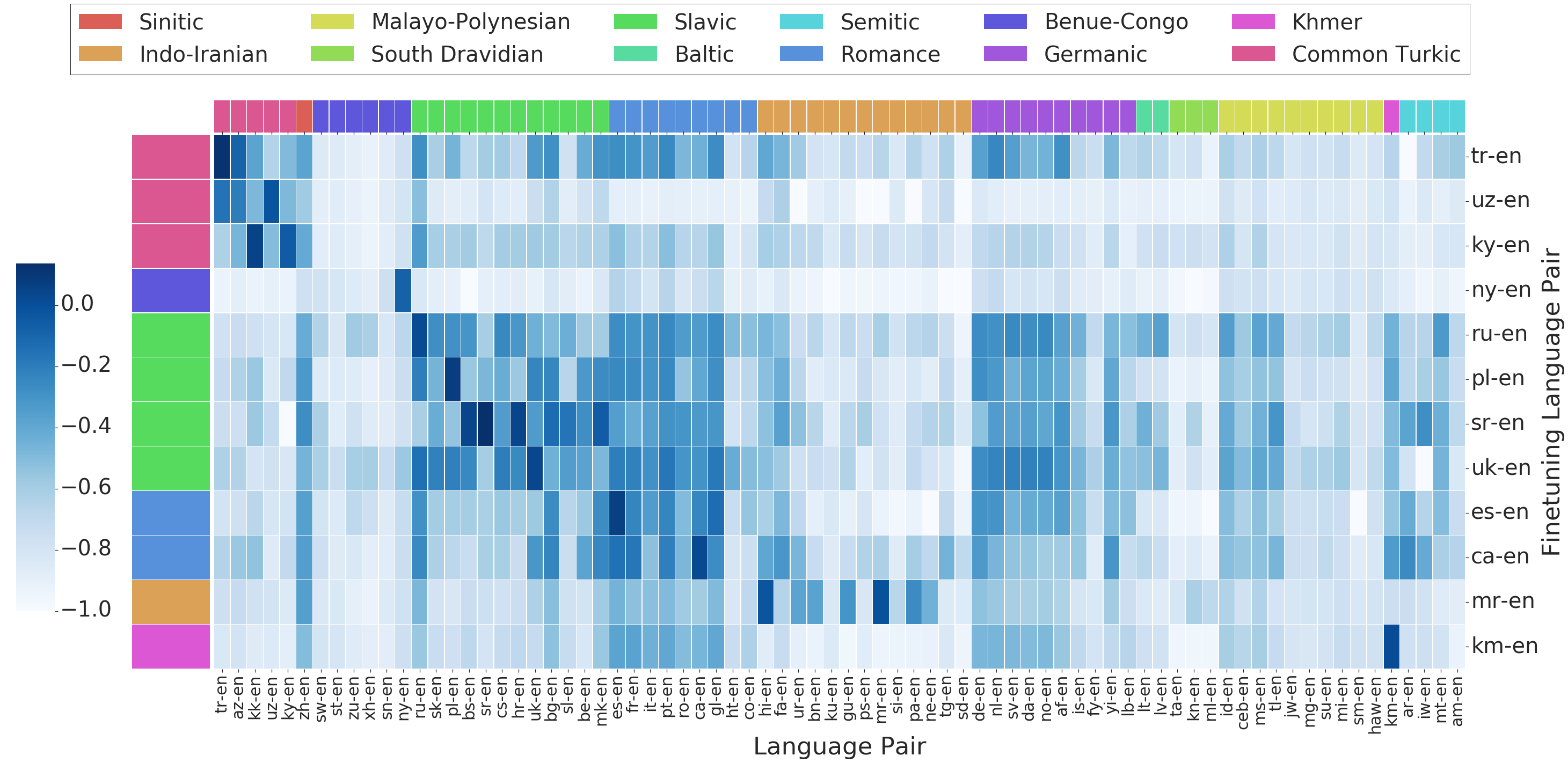}
\caption{Relative change in BLEU scores of \textit{xx-en} language pairs after finetuning with various language pairs of the form X-En. Darker cell means less change in BLEU score of \textit{xx-en} on finetuning with X-En.}
\label{fig:ft-bleu}
\end{subfigure}
\caption{Visualization depicting the (a) change in representations (using SVCCA) and (b) relative change in performance (in terms of test BLEU) of \textit{xx-en} language pairs (x-axis), after fine-tuning a large multilingual model on various X-En language pairs (y-axis). Language sub-families on the axes are color-coded. We notice that representations of high resource languages are relatively robust to fine-tuning on any language. Additionally, languages linguistically similar to the one being fine-tuned on observe less distortion from their original representations.}
\end{figure*}

\subsection{Representational Similarity evolves across Layers}\label{sub:layervar}

To visualize the how the representational overlap across languages evolves in the model, we plot how the distribution of pairwise SVCCA scores change across layers. For each layer, we first compute the pair-wise similarity between all pairs of languages. These similarity scores are then aggregated into a distribution and represented in Figures \ref{fig:xen-var} and \ref{fig:enx-var}, separately for the Any-to-English (X-En) and English-to-Any (En-X) language pairs.

For the Any-to-English (X-En) language pairs (Figure \ref{fig:xen-var}), we notice that similarity between the source languages (X) increase as we move up the encoder, from embeddings towards higher level encoder layers, suggesting that the encoder attempts to learn a common representation for all source languages. This might also be motivated by training on a shared target language (En). However, representations at the top of the encoder are far from perfectly aligned, possibly indicating that different languages are represented in only partially overlapping sub-spaces. On the other hand, as the decoder incorporates more information from the source language (X), representations of the target (En) diverge. This is also in line with some findings of studies on translationese \cite{koppel2011translationese}, where the authors show that that the translated text is predictive of the source language. For English-to-Any (En-X) language pairs (Figure \ref{fig:enx-var}) we observe a similar trend. Representations of the source language (En) diverge as we move up the encoder, indicating that the representations of English sentences separate conditioned on the target language. 

While it is a natural assumption that the encoder in a seq2seq model encodes the source, and the decoder decodes it into the target \cite{sutskever2014sequence,cho2014learning,BahdanauCB15}, our results indicate that this change from source to target might be more gradual, and the boundary between encoder and decoder, in terms of the localization of the representation is blurry.

\section{Analyzing representation robustness to fine-tuning}\label{sect:ft}

In this section, we try to analyze the robustness of encoder representations, when fine-tuning a multilingual model on a single language pair. Note that here we define robustness to mean representational rigidity to fine-tuning, aka robustness to catastrophic forgetting \cite{mccloskey1989catastrophic}. Understanding the factors that affect robustness during fine-tuning is critical to determining how much cross-lingual transfer can be expected for individual languages in a zero or few-shot setting.

\textbf{Analysis Setup}: We fine-tune a fully trained multilingual model separately on 12 Any-to-English language pairs for 40k steps using the optimizer as described in \ref{sec:model_details}. These selected languages form a mix of high and low resource language pairs, from 6 language sub-families.\footnote{More details on the relative resource sizes of different language pairs can be found in the Appendix \ref{sub:ft-size}.}

We first attempt to quantify the extent of distortion in language representations caused by the fine-tuning process. To this end, we calculate the SVCCA similarity between the encoder representations of a language, before and after fine-tuning. We do this for all languages, in order to understand which factors determine the extent of distortion. We visualize these changes in Figure \ref{fig:ft-xen} for the final encoder layer, for all X-En language pairs. To complement our analysis of representational similarity, we visualize the relative change in BLEU score after fine-tuning in Figure \ref{fig:ft-bleu}. 

\textbf{Observations}: The first observation from Figures \ref{fig:ft-xen} and \ref{fig:ft-bleu} is that the variations in SVCCA scores correspond very well with changes in BLEU; degradation in translation quality is strongly correlated with the magnitude of change in representations during fine-tuning. 

We find that representations of high resource languages are quite robust to fine-tuning on any language. In Figure \ref{fig:ft-xen}, we see that high resource languages such as Chinese, German, Russian and Italian do not change much, irrespective of the language the model is fine-tuned on. 

In general, we find that language representations are relatively robust to fine-tuning on a language pair from the same linguistic family. For example, on fine-tuning with tr-en (Turkish) or ky-en (Kyrgyz), the Turkic language group does not experience much shift in representation. We see a similar pattern with models fine-tuned on es-en (Spanish), ca-en (Catalan) and the Romance languages, uk-en (Ukranian), sr-en (Serbian), ru-en (Russian) and the Slavic languages.

An exception to these general trends seems to be fine-tuning on ny-en (Nyanja: Benue-Congo sub-family); all language pairs degrade by roughly the same extent, irrespective of language similarity or resource size. It's worth noting that all of the languages from the Benue-Congo sub-family are low-resource in our corpus.

These observations suggest that high resource languages might be responsible for partitioning the representation space, while low-resource languages become closely intertwined with linguistically similar high-resource languages. Low resource languages unrelated to any high resource languages have representations spread out in the subspace. 

While these observations are based on representations from the top of the encoder, we further analyze sensitivity of representations to fine-tuning across different layers in the Appendix \ref{sub:ft-var}. One key observation from that analysis is the robustness of embeddings to fine-tuning on any individual language; there is no significant change in the embedding representations (Correlation between representation of any language before and after finetuning $\bar{\rho} > 0.98$). 

\section{Discussion}\label{sect:impl}
Our work uncovers a few observations that might be of interest to practitioners working in multilingual NMT and cross-lingual transfer.

Our analysis reveals that language representations cluster based on language similarity. While language similarity has been exploited for adaptation previously \cite{neubig2018rapid}, our work is the first to concretely highlight which factors affect the overlap in representations across languages. This has potential implications for transfer learning: for example, it is possible to identify and exploit the nearest neighbors of a low resource language to maximize adaptation performance.

We also highlight how representation overlap evolves across layers, which is, again, of interest for cross-lingual transfer. For example, our analysis reveals that embeddings of different languages are less overlapping than the final encoder outputs. This hints that it might not be might not be effective to utilize input embeddings learned in multilingual NMT, since they don't overlap as much as the final encoder outputs. We also notice that encoder representation overlap across languages is not perfect, which explains why explicit language alignment or consistency losses might be needed to enable zero-shot NMT \cite{arivazhagan2019missing,DBLP:journals/corr/abs-1904-02338}.

We further analyze the robustness of language representations to fine-tuning, and notice that high-resource and linguistically similar languages are more robust to fine-tuning on an arbitrary language. This might help explain why linguistically distant languages typically result in poor zero-shot transfer \cite{DBLP:journals/corr/abs-1904-09077}. Applying explicit losses, like elastic-weight consolidation \cite{kirkpatrick2017overcoming}, to force language representations of distant languages from getting distorted might help improve transfer performance.

\section{Conclusion}\label{sect:discussion}

To conclude, we analyzed factors that affect the overlap in representations learned by multilingual NMT models. We used SVCCA to show that multilingual neural networks share representations across languages strongly along the lines of linguistic similarity, and encoder representations diverge based on the target language. With this work we hope to inspire future work on understanding multitask and multilingual NLP models.

\section*{Acknowledgments}

  We would like to thank Aditya Siddhant, Dmitry (Dima) Lepikhin, Klaus Macherey and Macduff Hughes for their helpful feedback on the draft. We would also like to thank Maithra Raghu for multiple insightful discussions about our methodology.

\bibliography{emnlp-ijcnlp-2019}

\begin{thebibliography}{50}
\expandafter\ifx\csname natexlab\endcsname\relax\def\natexlab#1{#1}\fi

\bibitem[{Aharoni et~al.(2019)Aharoni, Johnson, and
  Firat}]{aharoni2019massively}
Roee Aharoni, Melvin Johnson, and Orhan Firat. 2019.
\newblock Massively multilingual neural machine translation.
\newblock \emph{arXiv preprint arXiv:1903.00089}.

\bibitem[{Al{-}Shedivat and Parikh(2019)}]{DBLP:journals/corr/abs-1904-02338}
Maruan Al{-}Shedivat and Ankur~P. Parikh. 2019.
\newblock \href {http://arxiv.org/abs/1904.02338} {Consistency by agreement in
  zero-shot neural machine translation}.
\newblock \emph{CoRR}, abs/1904.02338.

\bibitem[{Arivazhagan et~al.(2019{\natexlab{a}})Arivazhagan, Bapna, Firat,
  Aharoni, Johnson, and Macherey}]{arivazhagan2019missing}
Naveen Arivazhagan, Ankur Bapna, Orhan Firat, Roee Aharoni, Melvin Johnson, and
  Wolfgang Macherey. 2019{\natexlab{a}}.
\newblock The missing ingredient in zero-shot neural machine translation.
\newblock \emph{arXiv preprint arXiv:1903.07091}.

\bibitem[{Arivazhagan et~al.(2019{\natexlab{b}})Arivazhagan, Bapna, Firat,
  Lepikhin, Johnson, Krikun, Chen, Cao, Foster, Cherry, Macherey, Chen, and
  Wu}]{arivazhagan2019massively}
Naveen Arivazhagan, Ankur Bapna, Orhan Firat, Dmitry Lepikhin, Melvin Johnson,
  Maxim Krikun, Mia~Xu Chen, Yuan Cao, George Foster, Colin Cherry, Wolfgang
  Macherey, Zhifeng Chen, and Yonghui Wu. 2019{\natexlab{b}}.
\newblock \href {http://arxiv.org/abs/1907.05019} {Massively multilingual
  neural machine translation in the wild: Findings and challenges}.

\bibitem[{Artetxe and Schwenk(2018)}]{artetxe2018massively}
Mikel Artetxe and Holger Schwenk. 2018.
\newblock Massively multilingual sentence embeddings for zero-shot
  cross-lingual transfer and beyond.
\newblock \emph{arXiv preprint arXiv:1812.10464}.

\bibitem[{Bahdanau et~al.(2015)Bahdanau, Cho, and Bengio}]{BahdanauCB15}
Dzmitry Bahdanau, Kyunghyun Cho, and Yoshua Bengio. 2015.
\newblock \href {http://arxiv.org/abs/1409.0473} {Neural machine translation by
  jointly learning to align and translate}.
\newblock In \emph{International Conference on Learning Representations}.

\bibitem[{Bau et~al.(2018)Bau, Belinkov, Sajjad, Durrani, Dalvi, and
  Glass}]{bau2018identifying}
Anthony Bau, Yonatan Belinkov, Hassan Sajjad, Nadir Durrani, Fahim Dalvi, and
  James Glass. 2018.
\newblock Identifying and controlling important neurons in neural machine
  translation.
\newblock \emph{arXiv preprint arXiv:1811.01157}.

\bibitem[{Belinkov et~al.(2017)Belinkov, Durrani, Dalvi, Sajjad, and
  Glass}]{belinkov2017neural}
Yonatan Belinkov, Nadir Durrani, Fahim Dalvi, Hassan Sajjad, and James Glass.
  2017.
\newblock What do neural machine translation models learn about morphology?
\newblock \emph{arXiv preprint arXiv:1704.03471}.

\bibitem[{Belinkov et~al.(2018)Belinkov, M{\`a}rquez, Sajjad, Durrani, Dalvi,
  and Glass}]{belinkov2018evaluating}
Yonatan Belinkov, Llu{\'\i}s M{\`a}rquez, Hassan Sajjad, Nadir Durrani, Fahim
  Dalvi, and James Glass. 2018.
\newblock Evaluating layers of representation in neural machine translation on
  part-of-speech and semantic tagging tasks.
\newblock \emph{arXiv preprint arXiv:1801.07772}.

\bibitem[{Belkin and Niyogi(2003)}]{belkin2003laplacian}
Mikhail Belkin and Partha Niyogi. 2003.
\newblock Laplacian eigenmaps for dimensionality reduction and data
  representation.
\newblock \emph{Neural computation}, 15(6):1373--1396.

\bibitem[{Browne and Ivanov()}]{slavtree}
Wayles Browne and Vyacheslav~Vsevolodovich Ivanov.
\newblock \href
  {https://www.britannica.com/topic/South-Slavic-languages/media/556411/61368}
  {{Slavic languages' family tree}}.
\newblock Encyclopædia Britannica.

\bibitem[{Chen et~al.(2018)Chen, Firat, Bapna, Johnson, Macherey, Foster,
  Jones, Schuster, Shazeer, Parmar, Vaswani, Uszkoreit, Kaiser, Chen, Wu, and
  Hughes}]{chen-EtAl:2018:Long1}
Mia~Xu Chen, Orhan Firat, Ankur Bapna, Melvin Johnson, Wolfgang Macherey,
  George Foster, Llion Jones, Mike Schuster, Noam Shazeer, Niki Parmar, Ashish
  Vaswani, Jakob Uszkoreit, Lukasz Kaiser, Zhifeng Chen, Yonghui Wu, and
  Macduff Hughes. 2018.
\newblock \href {http://www.aclweb.org/anthology/P18-1008} {The best of both
  worlds: Combining recent advances in neural machine translation}.
\newblock In \emph{Proceedings of the 56th Annual Meeting of the Association
  for Computational Linguistics (Volume 1: Long Papers)}, pages 76--86,
  Melbourne, Australia. Association for Computational Linguistics.

\bibitem[{Cho et~al.(2014)Cho, Van~Merri{\"e}nboer, Gulcehre, Bahdanau,
  Bougares, Schwenk, and Bengio}]{cho2014learning}
Kyunghyun Cho, Bart Van~Merri{\"e}nboer, Caglar Gulcehre, Dzmitry Bahdanau,
  Fethi Bougares, Holger Schwenk, and Yoshua Bengio. 2014.
\newblock Learning phrase representations using rnn encoder-decoder for
  statistical machine translation.
\newblock \emph{arXiv preprint arXiv:1406.1078}.

\bibitem[{Coperahawa(2007)}]{sinhalatamil}
Sandagomi Coperahawa. 2007.
\newblock {Language Contact and Linguistic Area: the Sinhala - Tamil Contact
  Situation}.
\newblock \emph{Journal of the Royal Asiatic Society of Sri Lanka}, 53:133 --
  152.

\bibitem[{Dalvi et~al.(2019)Dalvi, Durrani, Sajjad, Belinkov, Bau, and
  Glass}]{dalvi2019one}
Fahim Dalvi, Nadir Durrani, Hassan Sajjad, Yonatan Belinkov, D~Anthony Bau, and
  James Glass. 2019.
\newblock What is one grain of sand in the desert? analyzing individual neurons
  in deep nlp models.
\newblock In \emph{Proceedings of the AAAI Conference on Artificial
  Intelligence (AAAI)}, volume~7.

\bibitem[{Eriguchi et~al.(2018)Eriguchi, Johnson, Firat, Kazawa, and
  Macherey}]{eriguchi2018zero}
Akiko Eriguchi, Melvin Johnson, Orhan Firat, Hideto Kazawa, and Wolfgang
  Macherey. 2018.
\newblock Zero-shot cross-lingual classification using multilingual neural
  machine translation.
\newblock \emph{arXiv preprint arXiv:1809.04686}.

\bibitem[{Firat et~al.(2016)Firat, Cho, and Bengio}]{firat2016multi}
Orhan Firat, Kyunghyun Cho, and Yoshua Bengio. 2016.
\newblock Multi-way, multilingual neural machine translation with a shared
  attention mechanism.
\newblock \emph{arXiv preprint arXiv:1601.01073}.

\bibitem[{Hammarstr{\" o}m et~al.(2017)Hammarstr{\" o}m, Forkel, and
  Haspelmath}]{hindiurdu3}
Harald Hammarstr{\" o}m, Robert Forkel, and Martin Haspelmath. 2017.
\newblock \emph{{Hindustani}}.

\bibitem[{Hardoon et~al.(2004)Hardoon, Szedmak, and
  Shawe-Taylor}]{hardoon2004canonical}
David~R Hardoon, Sandor Szedmak, and John Shawe-Taylor. 2004.
\newblock Canonical correlation analysis: An overview with application to
  learning methods.
\newblock \emph{Neural computation}, 16(12):2639--2664.

\bibitem[{Herzog()}]{yiddish}
Marvin~Irving Herzog.
\newblock \href
  {https://www.britannica.com/topic/West-Germanic-languages/German#ref603830}
  {{West Germanic Languages}}.

\bibitem[{Ivi{\' c}(2011)}]{brittanicalanguageclassification}
Pavle Ivi{\' c}. 2011.
\newblock \href
  {https://www.britannica.com/science/linguistics/Language-classification}
  {{Encyclopaedia Britannica}}.

\bibitem[{Johnson et~al.(2017)Johnson, Schuster, Le, Krikun, Wu, Chen, Thorat,
  Vi{\'e}gas, Wattenberg, Corrado et~al.}]{johnson2017google}
Melvin Johnson, Mike Schuster, Quoc~V Le, Maxim Krikun, Yonghui Wu, Zhifeng
  Chen, Nikhil Thorat, Fernanda Vi{\'e}gas, Martin Wattenberg, Greg Corrado,
  et~al. 2017.
\newblock Google’s multilingual neural machine translation system: Enabling
  zero-shot translation.
\newblock \emph{Transactions of the Association for Computational Linguistics},
  5:339--351.

\bibitem[{Kirkpatrick et~al.(2017)Kirkpatrick, Pascanu, Rabinowitz, Veness,
  Desjardins, Rusu, Milan, Quan, Ramalho, Grabska-Barwinska
  et~al.}]{kirkpatrick2017overcoming}
James Kirkpatrick, Razvan Pascanu, Neil Rabinowitz, Joel Veness, Guillaume
  Desjardins, Andrei~A Rusu, Kieran Milan, John Quan, Tiago Ramalho, Agnieszka
  Grabska-Barwinska, et~al. 2017.
\newblock Overcoming catastrophic forgetting in neural networks.
\newblock \emph{Proceedings of the national academy of sciences},
  114(13):3521--3526.

\bibitem[{Koppel and Ordan(2011)}]{koppel2011translationese}
Moshe Koppel and Noam Ordan. 2011.
\newblock Translationese and its dialects.
\newblock In \emph{Proceedings of the 49th Annual Meeting of the Association
  for Computational Linguistics: Human Language Technologies-Volume 1}, pages
  1318--1326. Association for Computational Linguistics.

\bibitem[{Kudo and Richardson(2018)}]{kudo2018sentencepiece}
Taku Kudo and John Richardson. 2018.
\newblock Sentencepiece: A simple and language independent subword tokenizer
  and detokenizer for neural text processing.
\newblock \emph{arXiv preprint arXiv:1808.06226}.

\bibitem[{Lample and Conneau(2019)}]{lample2019cross}
Guillaume Lample and Alexis Conneau. 2019.
\newblock Cross-lingual language model pretraining.
\newblock \emph{arXiv preprint arXiv:1901.07291}.

\bibitem[{McCloskey and Cohen(1989)}]{mccloskey1989catastrophic}
Michael McCloskey and Neal~J Cohen. 1989.
\newblock Catastrophic interference in connectionist networks: The sequential
  learning problem.
\newblock In \emph{Psychology of learning and motivation}, volume~24, pages
  109--165. Elsevier.

\bibitem[{Moorti(2011)}]{protodravidian}
Etukoori~Balaraama Moorti. 2011.
\newblock \href
  {https://lists.hcs.harvard.edu/mailman/listinfo/proto-dravidian} {{Proto
  Dravidian}}.
\newblock \emph{Study of Dravidian Linguistics and Civilization}.

\bibitem[{Morcos et~al.(2018)Morcos, Raghu, and
  Bengio}]{Morcos:2018:IRS:3327345.3327475}
Ari~S. Morcos, Maithra Raghu, and Samy Bengio. 2018.
\newblock \href {http://dl.acm.org/citation.cfm?id=3327345.3327475} {Insights
  on representational similarity in neural networks with canonical
  correlation}.
\newblock In \emph{Proceedings of the 32Nd International Conference on Neural
  Information Processing Systems}, NIPS'18, pages 5732--5741, USA. Curran
  Associates Inc.

\bibitem[{Neubig and Hu(2018)}]{neubig2018rapid}
Graham Neubig and Junjie Hu. 2018.
\newblock Rapid adaptation of neural machine translation to new languages.
\newblock \emph{arXiv preprint arXiv:1808.04189}.

\bibitem[{Nguyen and Chiang(2017)}]{nguyen-chiang:2017:I17-2}
Toan~Q. Nguyen and David Chiang. 2017.
\newblock Transfer learning across low-resource, related languages for neural
  machine translation.
\newblock In \emph{Proc. IJCNLP}, volume~2, pages 296--301.

\bibitem[{Olah et~al.(2018)Olah, Satyanarayan, Johnson, Carter, Schubert, Ye,
  and Mordvintsev}]{olah2018building}
Chris Olah, Arvind Satyanarayan, Ian Johnson, Shan Carter, Ludwig Schubert,
  Katherine Ye, and Alexander Mordvintsev. 2018.
\newblock The building blocks of interpretability.
\newblock \emph{Distill}, 3(3):e10.

\bibitem[{{\"O}stling and Tiedemann(2016)}]{ostling2016continuous}
Robert {\"O}stling and J{\"o}rg Tiedemann. 2016.
\newblock Continuous multilinguality with language vectors.
\newblock \emph{arXiv preprint arXiv:1612.07486}.

\bibitem[{Phillips and Davis(2009)}]{bcp47}
A.~Phillips and M~Davis. 2009.
\newblock \href {https://tools.ietf.org/html/bcp47} {{Tags for Identifying
  Languages}}.
\newblock {RFC} 5646, RFC Editor.

\bibitem[{Raganato and Tiedemann(2018)}]{raganato2018analysis}
Alessandro Raganato and J{\"o}rg Tiedemann. 2018.
\newblock An analysis of encoder representations in transformer-based machine
  translation.
\newblock In \emph{Proceedings of the 2018 EMNLP Workshop BlackboxNLP:
  Analyzing and Interpreting Neural Networks for NLP}, pages 287--297.

\bibitem[{Raghu et~al.(2017)Raghu, Gilmer, Yosinski, and
  Sohl-Dickstein}]{raghu2017svcca}
Maithra Raghu, Justin Gilmer, Jason Yosinski, and Jascha Sohl-Dickstein. 2017.
\newblock Svcca: Singular vector canonical correlation analysis for deep
  learning dynamics and interpretability.
\newblock In \emph{Advances in Neural Information Processing Systems}, pages
  6076--6085.

\bibitem[{Sahbi(2018)}]{sahbi2018learning}
Hichem Sahbi. 2018.
\newblock Learning cca representations for misaligned data.
\newblock In \emph{Proceedings of the European Conference on Computer Vision
  (ECCV)}, pages 0--0.

\bibitem[{Saphra and Lopez(2018)}]{saphra2018understanding}
Naomi Saphra and Adam Lopez. 2018.
\newblock Understanding learning dynamics of language models with svcca.
\newblock \emph{arXiv preprint arXiv:1811.00225}.

\bibitem[{Shazeer and Stern(2018)}]{shazeer2018adafactor}
Noam Shazeer and Mitchell Stern. 2018.
\newblock Adafactor: Adaptive learning rates with sublinear memory cost.
\newblock \emph{arXiv preprint arXiv:1804.04235}.

\bibitem[{Siddiqi(1994)}]{hindiurdu1}
Mohammad~Tahsin Siddiqi. 1994.
\newblock \emph{{ Hindustani-English code-mixing in modern literary texts}}.
\newblock University of Wisconsin.

\bibitem[{Sussex and Cubberley()}]{slaviclangs}
Roland Sussex and Paul Cubberley.
\newblock \href
  {http://assets.cambridge.org/97805212/94485/excerpt/9780521294485_excerpt.pdf}
  {{The Slavic Languages}}.

\bibitem[{Sutskever et~al.(2014)Sutskever, Vinyals, and
  Le}]{sutskever2014sequence}
Ilya Sutskever, Oriol Vinyals, and Quoc~V. Le. 2014.
\newblock Sequence to sequence learning with neural networks.
\newblock In \emph{Advances in Neural Information Processing Systems}, pages
  3104--3112.

\bibitem[{Tan et~al.(2019)Tan, Chen, He, Xia, Qin, and
  Liu}]{tan2019multilingual}
Xu~Tan, Jiale Chen, Di~He, Yingce Xia, Tao Qin, and Tie-Yan Liu. 2019.
\newblock Multilingual neural machine translation with language clustering.
\newblock \emph{arXiv preprint arXiv:1908.09324}.

\bibitem[{Tenney et~al.(2019)Tenney, Xia, Chen, Wang, Poliak, McCoy, Kim,
  Van~Durme, Bowman, Das et~al.}]{tenney2019you}
Ian Tenney, Patrick Xia, Berlin Chen, Alex Wang, Adam Poliak, R~Thomas McCoy,
  Najoung Kim, Benjamin Van~Durme, Samuel~R Bowman, Dipanjan Das, et~al. 2019.
\newblock What do you learn from context? probing for sentence structure in
  contextualized word representations.
\newblock \emph{arXiv preprint arXiv:1905.06316}.

\bibitem[{Tiedemann(2018)}]{tiedemann2018emerging}
J{\"o}rg Tiedemann. 2018.
\newblock Emerging language spaces learned from massively multilingual corpora.
\newblock \emph{arXiv preprint arXiv:1802.00273}.

\bibitem[{Uszkoreit et~al.(2010)Uszkoreit, Ponte, Popat, and
  Dubiner}]{uszkoreit2010large}
Jakob Uszkoreit, Jay~M Ponte, Ashok~C Popat, and Moshe Dubiner. 2010.
\newblock Large scale parallel document mining for machine translation.
\newblock In \emph{Proceedings of the 23rd International Conference on
  Computational Linguistics}, pages 1101--1109. Association for Computational
  Linguistics.

\bibitem[{Vaswani et~al.(2017)Vaswani, Shazeer, Parmar, Uszkoreit, Jones,
  Gomez, Kaiser, and Polosukhin}]{vaswani2017attention}
Ashish Vaswani, Noam Shazeer, Niki Parmar, Jakob Uszkoreit, Llion Jones,
  Aidan~N Gomez, {\L}ukasz Kaiser, and Illia Polosukhin. 2017.
\newblock Attention is all you need.
\newblock In \emph{Advances in Neural Information Processing Systems}, pages
  5998--6008.

\bibitem[{Wu and Dredze(2019)}]{DBLP:journals/corr/abs-1904-09077}
Shijie Wu and Mark Dredze. 2019.
\newblock \href {http://arxiv.org/abs/1904.09077} {Beto, bentz, becas: The
  surprising cross-lingual effectiveness of {BERT}}.
\newblock \emph{CoRR}, abs/1904.09077.

\bibitem[{Zhang and Bowman(2018)}]{zhang2018language}
Kelly~W Zhang and Samuel~R Bowman. 2018.
\newblock Language modeling teaches you more syntax than translation does:
  Lessons learned through auxiliary task analysis.
\newblock \emph{arXiv preprint arXiv:1809.10040}.

\bibitem[{Zoph et~al.(2016)Zoph, Yuret, May, and Knight}]{zoph2016transfer}
Barret Zoph, Deniz Yuret, Jonathan May, and Kevin Knight. 2016.
\newblock Transfer learning for low-resource neural machine translation.
\newblock In \emph{Proceedings of the 2016 Conference on Empirical Methods in
  Natural Language Processing}, pages 1568--1575.

\end{thebibliography}
\bibliographystyle{acl_natbib}

\appendix

\clearpage

\section{Supplemental Material}
\label{sec:supplemental}

\subsection{Model and Training Details}
\label{sec:model_details}

For all of the bilingual baselines and multilingual model that we investigate, we use the Transformer \cite{vaswani2017attention} architecture. In particular, we use the Transformer Big model containing 375M parameters in \citep{chen-EtAl:2018:Long1}. For multilingual models, we share all parameters across language pairs including softmax layer in input/output word embeddings. 

During training, we use a temperature based data sampling strategy, similar to the strategy used to train the multilingual models in \cite{arivazhagan2019massively}. That is, if $p_L$ is the probability that a sentence in the corpus belongs to language pair $L$, we sample from a distribution where the probability of sampling from $L$ is proportional to ${p_L}^{\frac{1}{T}}$. All the experiments in this paper are performed on a model trained with a sampling temperature $T=5$. 

For the vocabulary, we use a Sentence Piece Model (SPM) \cite{kudo2018sentencepiece} with 64k tokens shared on both the encoder and decoder side. To learn a joint SPM model given our imbalanced dataset, we followed the temperature based sampling strategy with a temperature of $T=5$.

Finally, all models are optimized using Adafactor optimizer \citep{shazeer2018adafactor} with momentum factorization and a per-parameter norm clipping threshold of 1.0. During optimization, we followed a learning rate of a learning rate of 3.0,  with 40K warm-up steps for the schedule,  which is decayed with the inverse square root of the number of training steps after warm-up. BLEU scores presented in this paper are calculated on true-cased output and references, where we used mteval-v13a.pl script from Moses.

\subsection{Baselines and Multilingual BLEU Scores}\label{sub:bleu_orig}

\begin{figure}[t!]
\begin{subfigure}{0.98\columnwidth}
  \centering
  \includegraphics[scale=0.225]{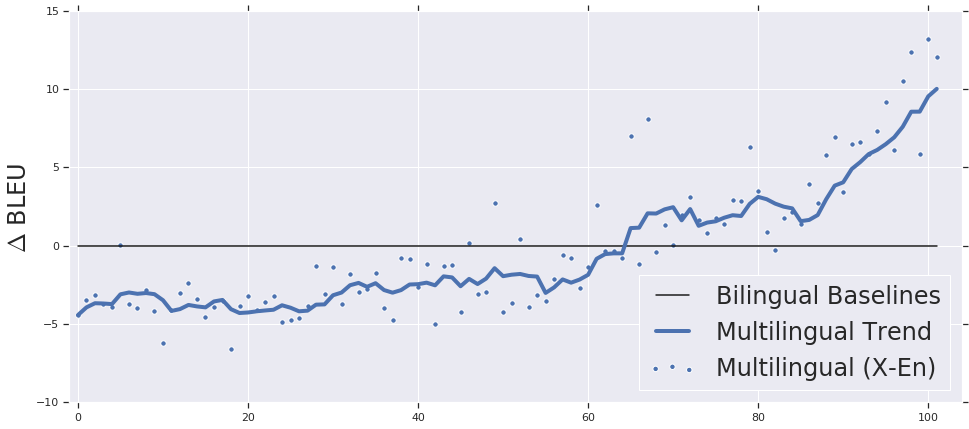}       
  \caption{Comparison of X-En pairs with baselines.}
\label{fig:baselines:x_en}
  \end{subfigure} 
    
\begin{subfigure}{0.98\columnwidth}
  \centering
  \includegraphics[scale=0.225]{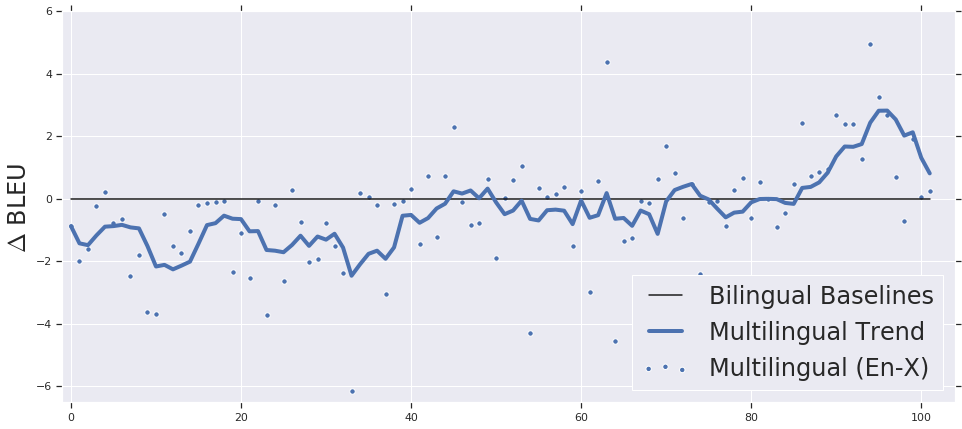}       
  \caption{Comparison of En-X pairs with baselines.}
\label{fig:baselines:en_x}
  \end{subfigure}
\caption{Trendlines depicting translation performance of the massively multilingual model (blue curves) compared to bilingual baselines (solid black lines). From left to right, languages are arranged in decreasing order of available training data (high-resource to low-resource). y-axis depicts the BLEU score relative to the bilingual baseline trained on the corresponding language pair. Top panel for Any-to-English pairs and bottom panel for English-to-Any pairs.}
\label{fig:baselines}
\end{figure}

To assess the quality of our single massively multilingual model trained on 103 languages, we trained bilingual baselines using the same training data, with models that are comparable in their size. For high resource languages, we trained identical architecture models (Transformer-Big) and only for a few low-resource languages we trained smaller models with heavy regularization (Transformer-Base). Results are shared in Figure~\ref{fig:baselines}. Note that, x-axes correspond to a different language pairs sorted with respect to the available training data and y-axes correspond to the divergence from the baseline BLEU scores. For each language pair, the BLEU scores are calculated on the test set that is specific for that language pair. From Figure~\ref{fig:baselines} it is clear that our massively multilingual model is dramatically better on low-resource languages (right-most portion of both panels) with some regression on high-resource languages (left-most portion of the panels). We provide the comparison with baselines to ground our analysis of massively multilingual model, which is competitive with bilingual baselines in quality.

\subsection{CCA for Misaligned Sequences}\label{sub:svcca-seq}

\begin{figure*}[ht!]
\centering
  \includegraphics[width=0.8\linewidth]{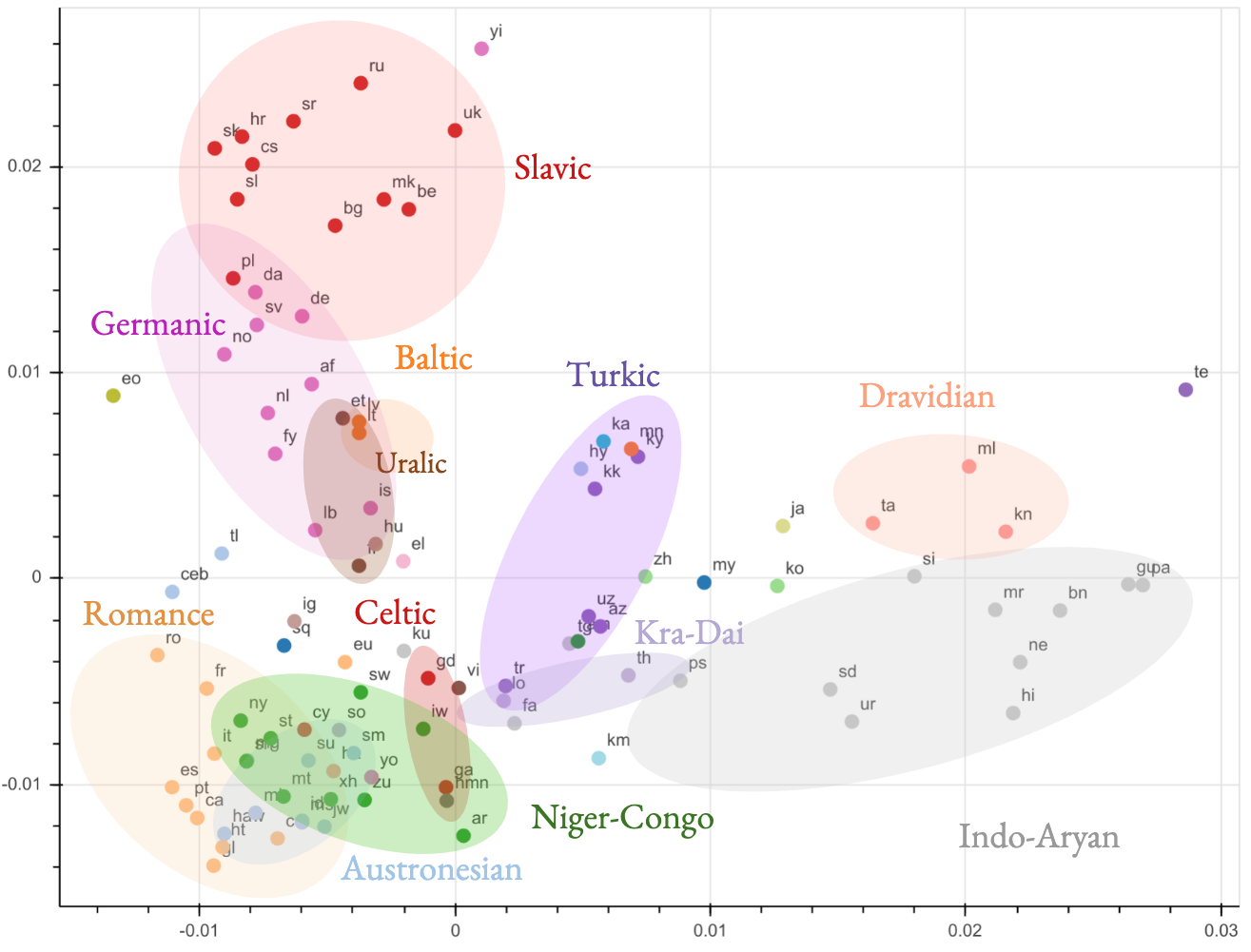}
  \caption{Top layer of the encoder for En-X language pairs using token level SVCCA as a similarity measure.}
  \label{fig:enxenctoklevel}
\end{figure*}

\begin{figure}[ht!]
\centering
  \includegraphics[width=\linewidth]{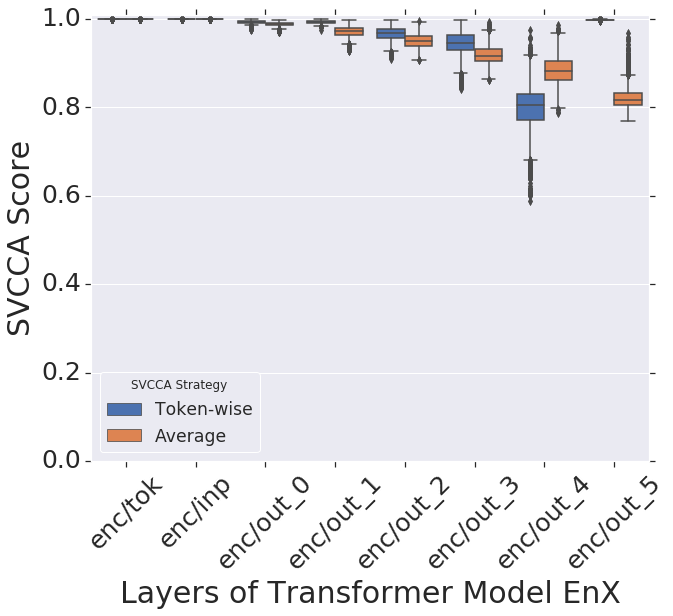}
  \caption{The change in distribution of pairwise SVCCA scores using our pooling strategy and a naive token-wise strategy between English-to-Any language pairs across the encoder layers of a multilingual NMT model. We see that while the encoder representations diverge in both cases, the top layer of the encoder does not seem to show any divergence for the token-wise strategy and is a possible artifact of the strategy.}
\label{fig:tok-evolve}
\end{figure}

In Section \ref{sub:svcca}, we discussed how the mean pooling strategy that we use is more suitable for our problem, where we use SVCCA to compare unaligned sequences. In this section, we attempt to replicate some results in the paper using a token-level CCA strategy, and discuss the differences in our results. In the mean pooling strategy,  each datapoint forming the subspace representing a language's encoding for a given layer is the mean of all timestep activations in a single sentence. On the other hand, in what we refer to as a token-level strategy, each data point is the activation of a timestep, with no differentiation between different sentences or positions.

\begin{figure*}[ht!]
\begin{subfigure}{\textwidth}
  \centering
  \includegraphics[width=0.8\linewidth]{lang_dist.png}       
  \caption{Top layer of the encoder for X-En language pairs.}
  \label{fig:xxenc}
  \end{subfigure}
\begin{subfigure}{\textwidth}
  \centering
  \includegraphics[width=0.8\linewidth]{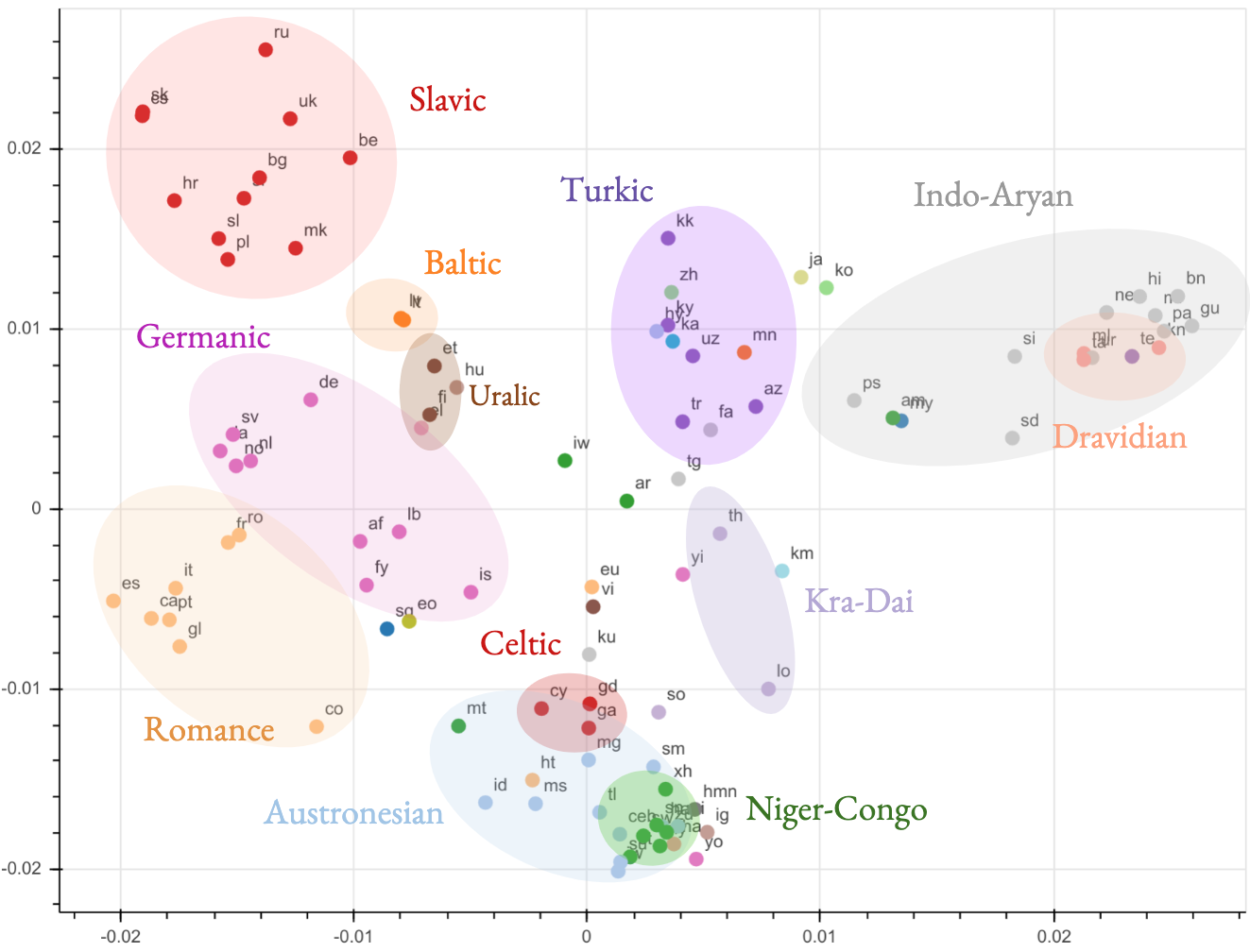}
  \caption{Top layer of the decoder for En-X language pairs.}
  \label{fig:xxdec}
  \end{subfigure}
\caption{Visualization of the top layer of the encoder and decoder. Both the encoder and decoder show clustering according to linguistic similarity.}
\label{fig:top-clusters}
\end{figure*}

In Figure \ref{fig:enxenctoklevel}, we plot the cluster formed by the SVCCA scores of English-to-Any language pairs using the method described in Section \ref{sub:vis}. Our data is unaligned for compared other components of our experiment, so we do not discuss those results. While we do see some amount of clustering according to linguistic similarity, the clusters are less separated than in Figure \ref{fig:top-clusters}. We also compare the distributions of pairwise SVCCA scores using our pooling strategy and a naive token-wise strategy between English-to-Any language pairs across layers of the encoder. We see similar trends upto the top layer of the encoder - this could possibly be an artifact of the naive strategy.

\subsection{Additional Clustering Visualization}\label{sub:extra-vis}

\begin{figure*}[ht!]
\begin{subfigure}{0.5\textwidth}
  \centering
  \includegraphics[width=.8\linewidth]{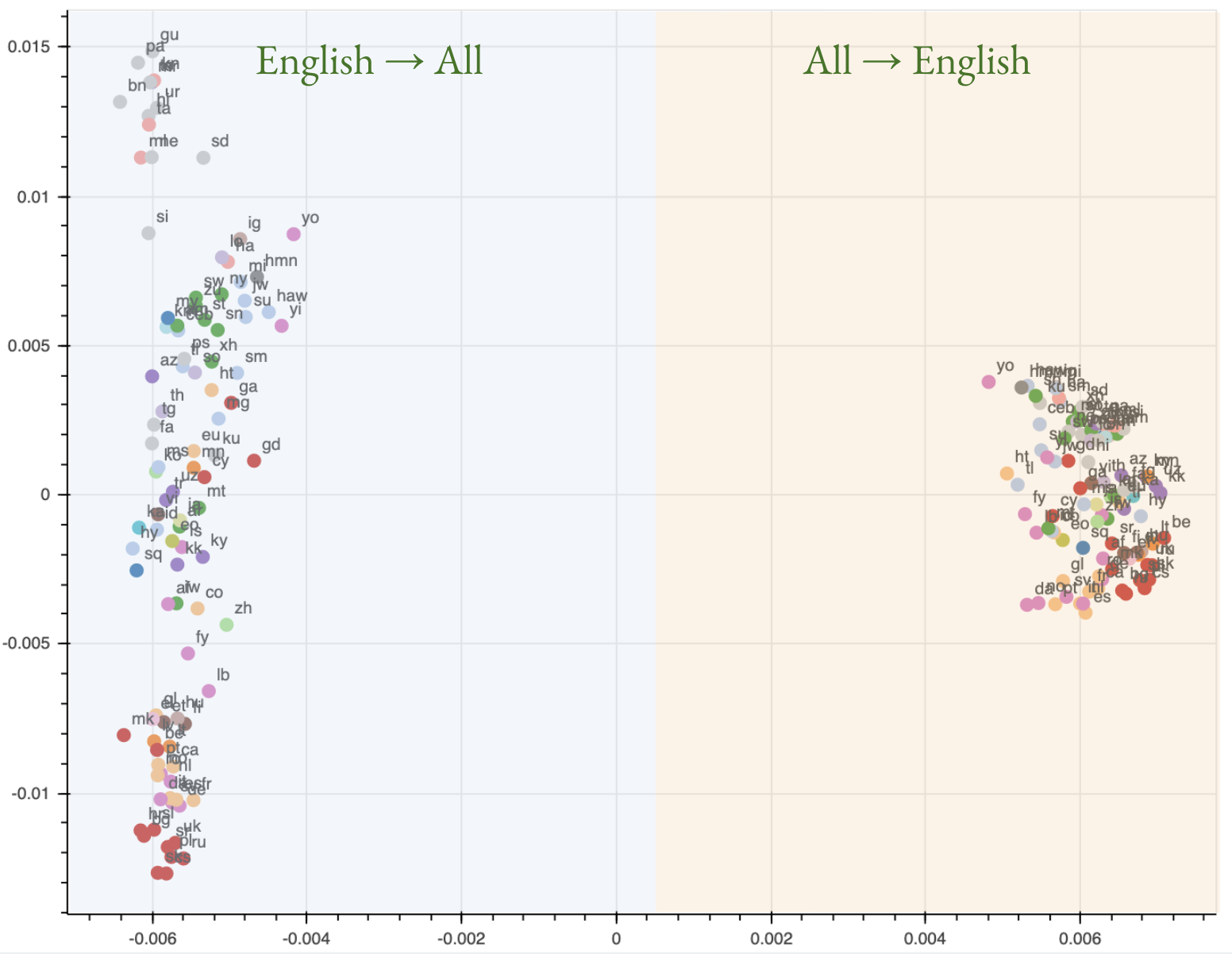}
  \caption{Top layer of the encoder.}
  \label{fig:xxenc}
  \end{subfigure}
\begin{subfigure}{0.5\textwidth}
  \centering
  \includegraphics[width=.8\linewidth]{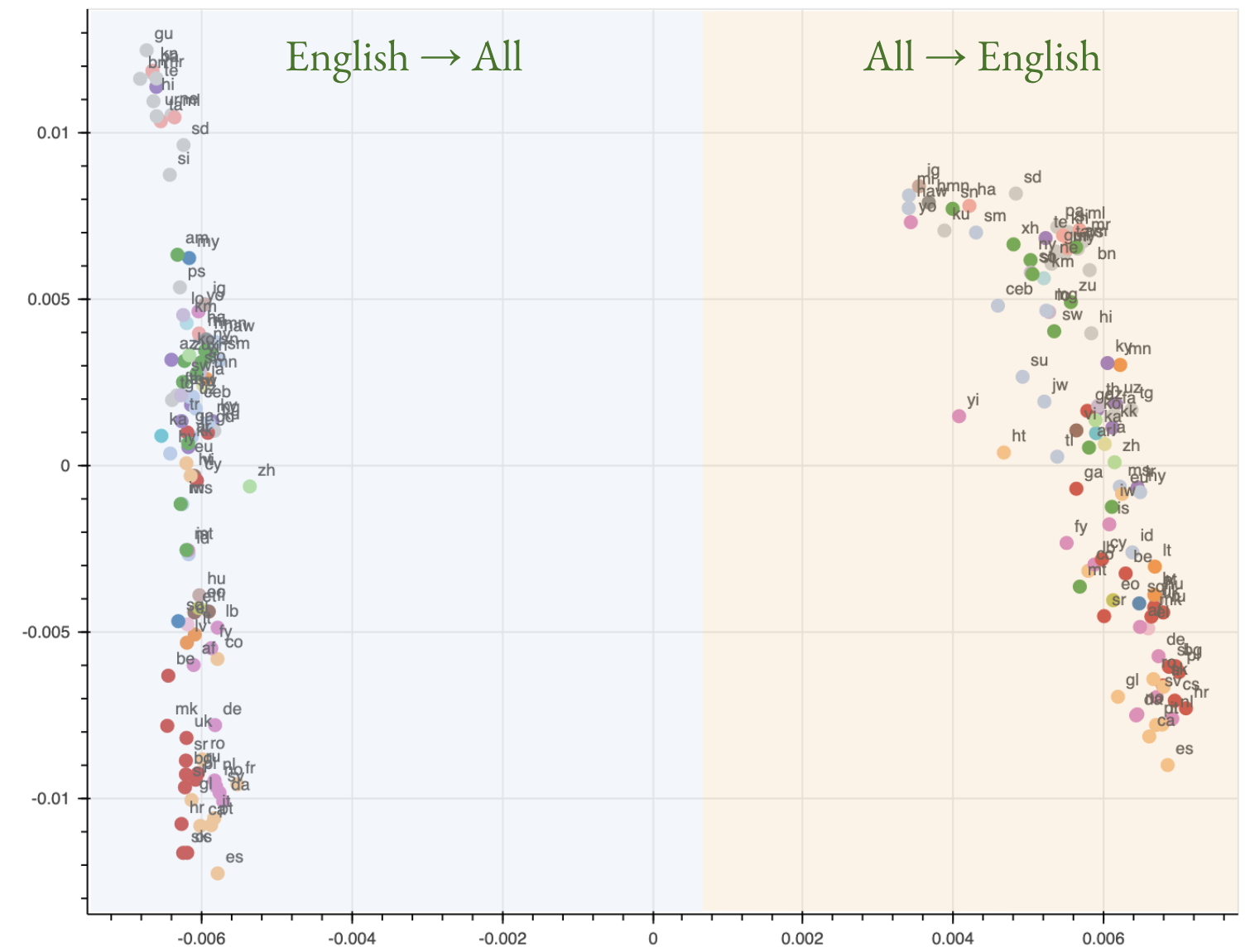}
  \caption{Top layer of the decoder.}
  \label{fig:xxdec}
  \end{subfigure}
\caption{All-to-All (X-X) clustering of the encoder and decoder representations of all languages, based on their SVCCA similarity on our multi-way parallel evaluation set.}
\label{fig:xx-clusters}

\begin{subfigure}{0.5\textwidth}
  \centering
  \includegraphics[width=.8\linewidth]{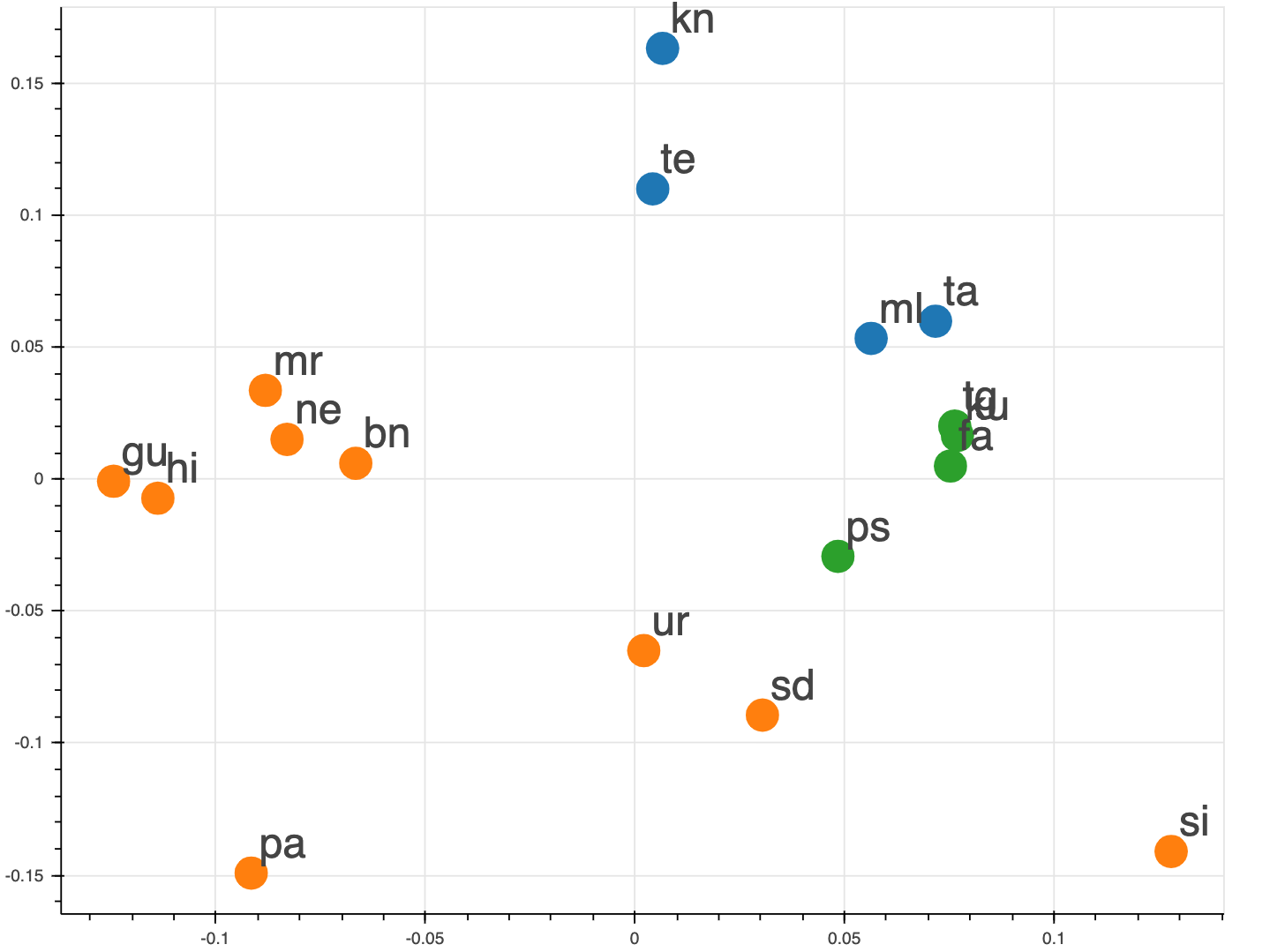}       
  \caption{Embeddings, colored by language group}
\label{fig:indo-iranian-dravidian-clusters:emb-fam}
  \end{subfigure}
\begin{subfigure}{0.5\textwidth}
  \centering
  \includegraphics[width=.8\linewidth]{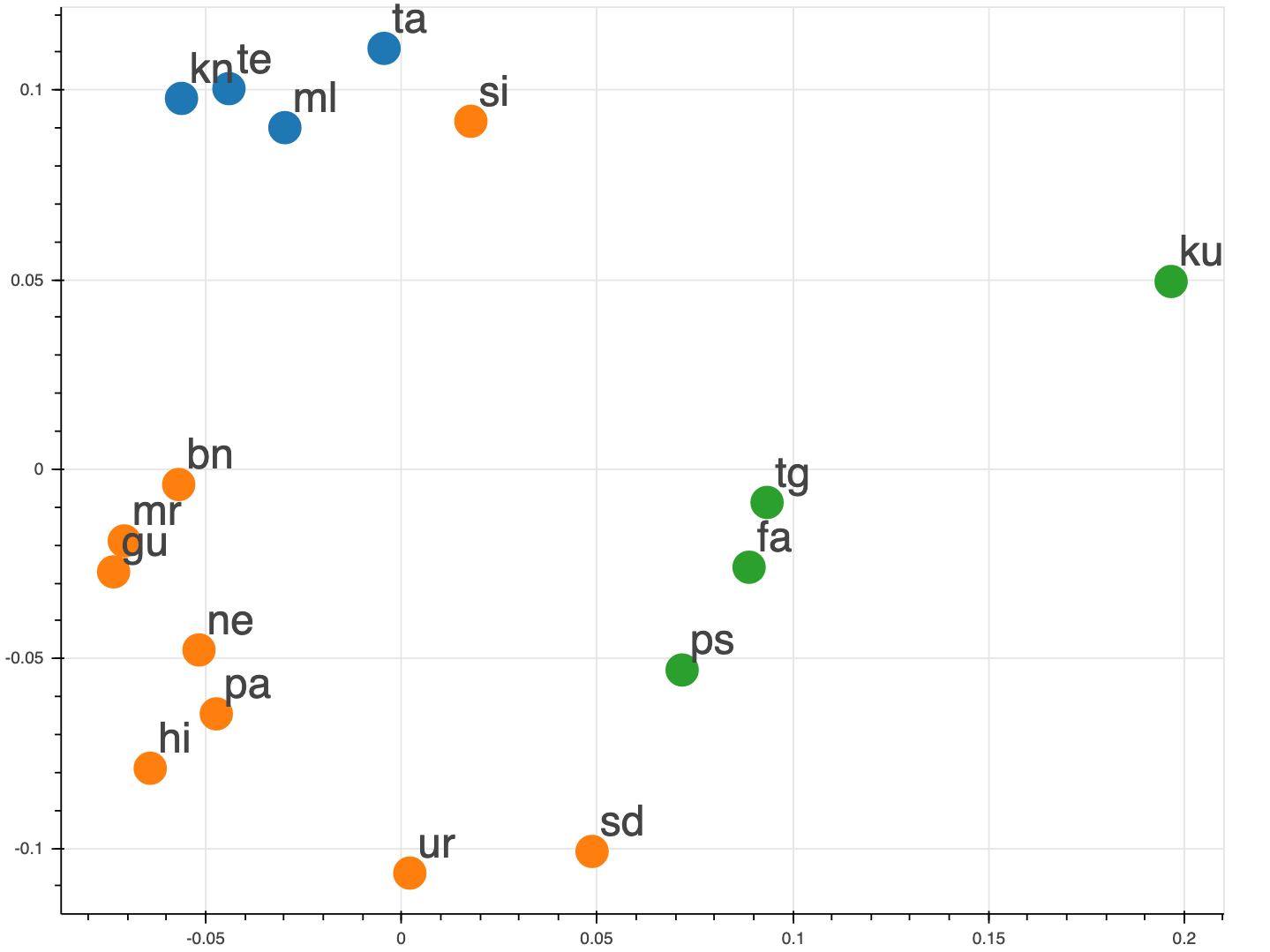}       
  \caption{Encoder layer 5, colored by language group}
\label{fig:indo-iranian-dravidian-clusters:enc5-fam}
  \end{subfigure}
\begin{subfigure}{0.5\textwidth}
  \centering
  \includegraphics[width=.8\linewidth]{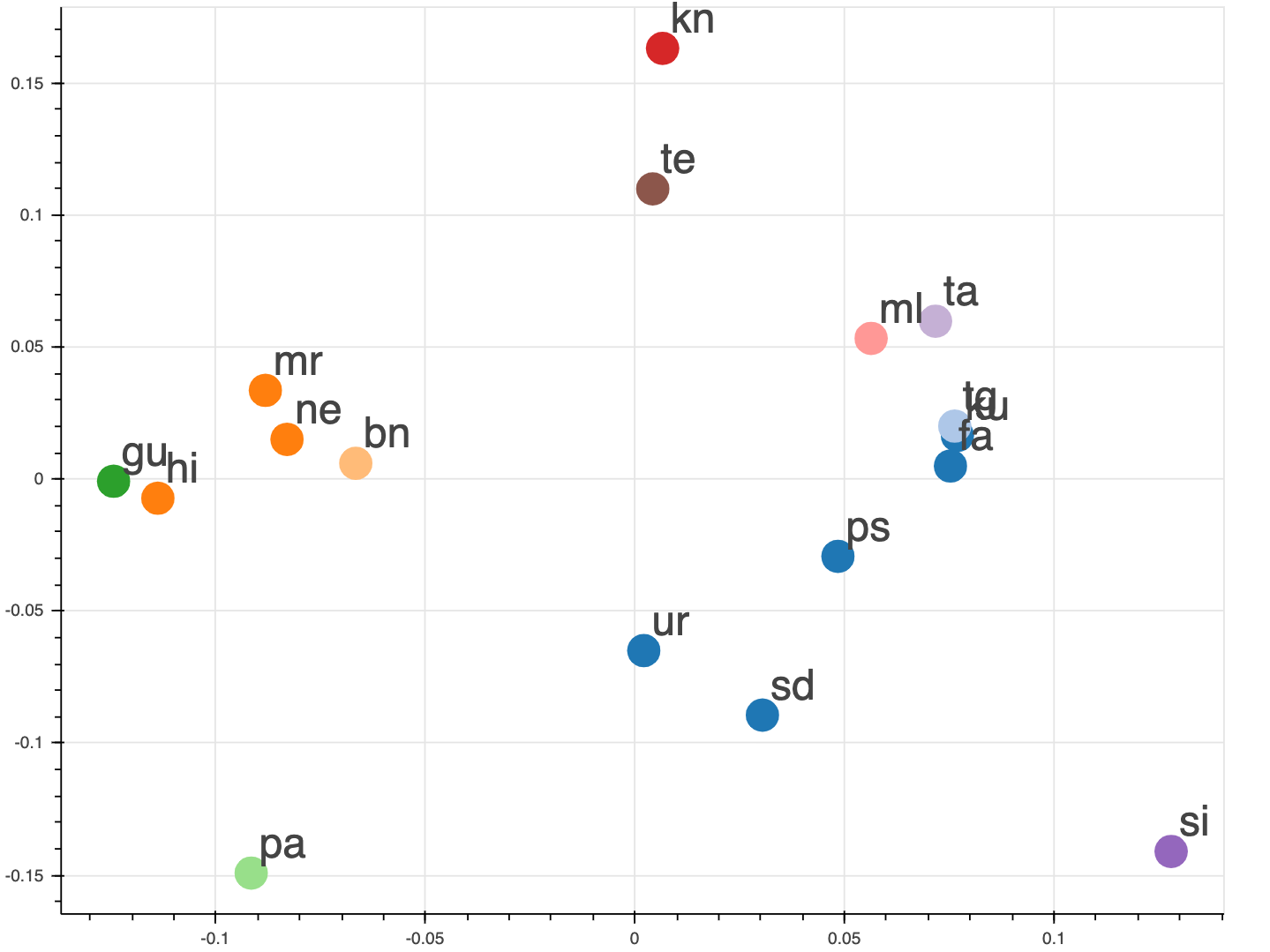}
  \caption{Embeddings, colored by script}
\label{fig:indo-iranian-dravidian-clusters:emb-script}
  \end{subfigure}
\begin{subfigure}{0.5\textwidth}
  \centering
  \includegraphics[width=.8\linewidth]{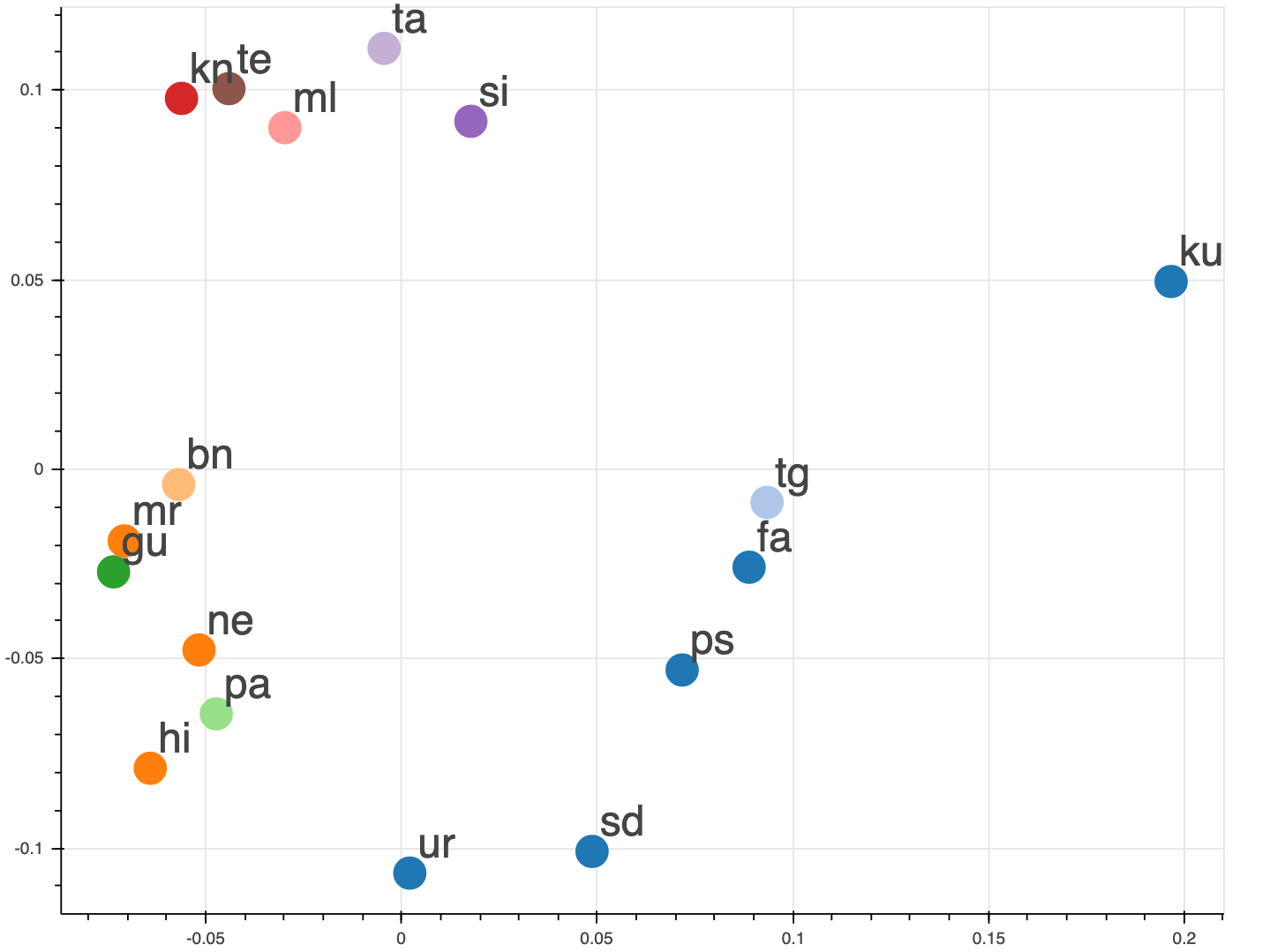}
  \caption{Encoder layer 5, colored by script}
\label{fig:indo-iranian-dravidian-clusters:enc5-script}
  \end{subfigure}
\caption{Visualization of the Indo-Aryan languages, the Iranian languages, and the Dravidian languages, for the embeddings (left column) and the top layer of the encoder (right column), coloring by different attributes to highlight clusters. These are to-English direction.}
\label{fig:indo-iranian-dravidian-clusters}
\end{figure*}

Here, we plot the clusters formed by all languages pairs, color coded by linguistic subfamily for the top layer of both the encoder and decoder. As seen in Figure \ref{fig:xx-clusters}, there is a clear separation between languages of the form En-X and X-En. So, we cluster the En-X and X-En language pairs (for the top layer of the decoder and encoder respectively) separately in Figure \ref{fig:xxenc} and Figure \ref{fig:xxdec}. The activations of the token embedding layers do not separate significantly, so we do not cluster them.

\subsubsection*{Low-resource, script-diverse language families}

\begin{figure*}[ht!]
\begin{subfigure}{0.5\textwidth}
  \centering
  \includegraphics[width=.8\linewidth]{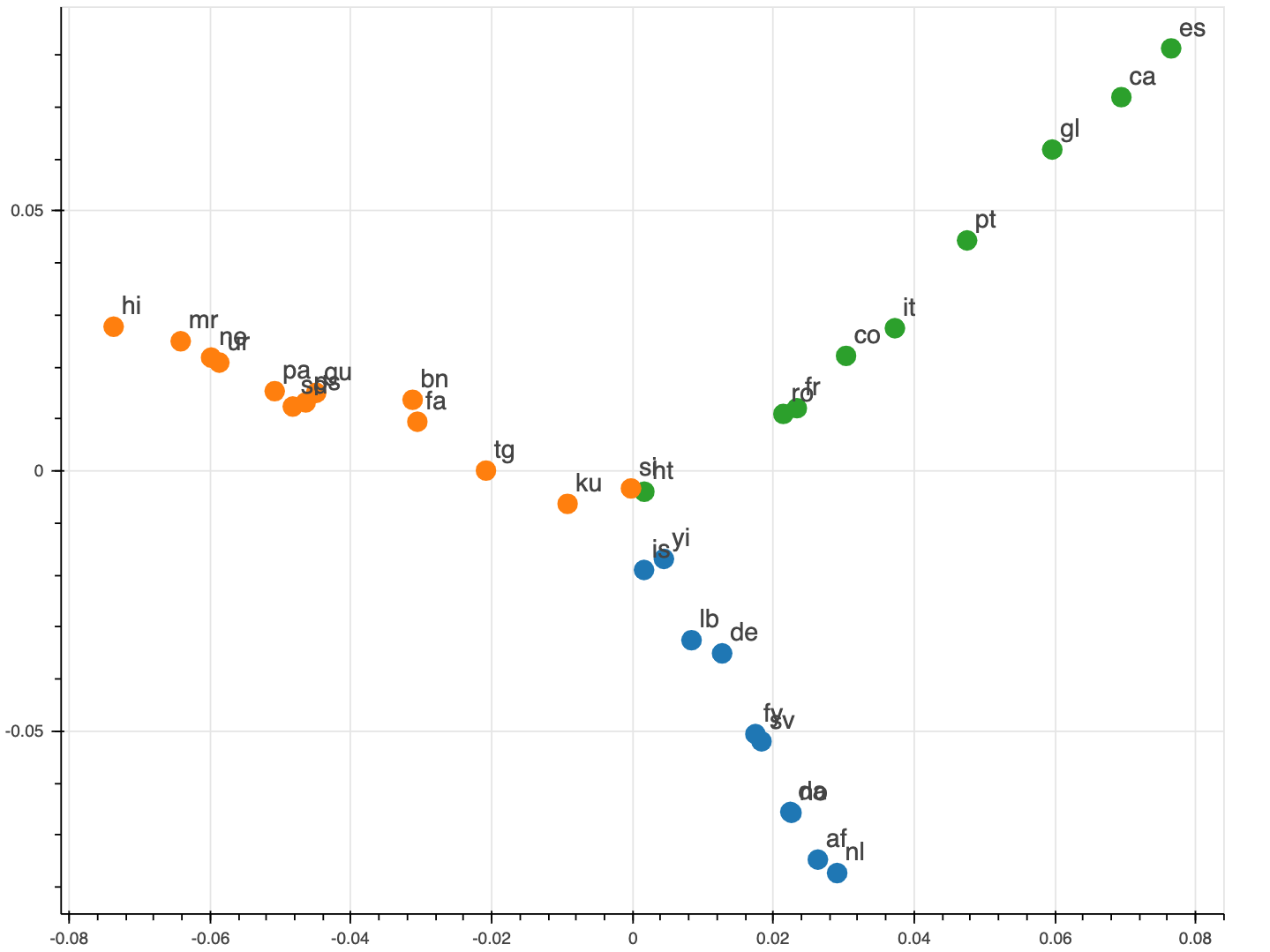}    
  \captionsetup{width=.8\linewidth}
  \caption{Colored by language group:  Indo-Iranian languages in red, Germanic languages in blue, and Romance languages in green. Note that within the Indo-Iranian languages, the Iranian languages are to the right, and the Indic languages to the left.}
  \end{subfigure}
\begin{subfigure}{0.5\textwidth}
  \centering
  \includegraphics[width=.8\linewidth]{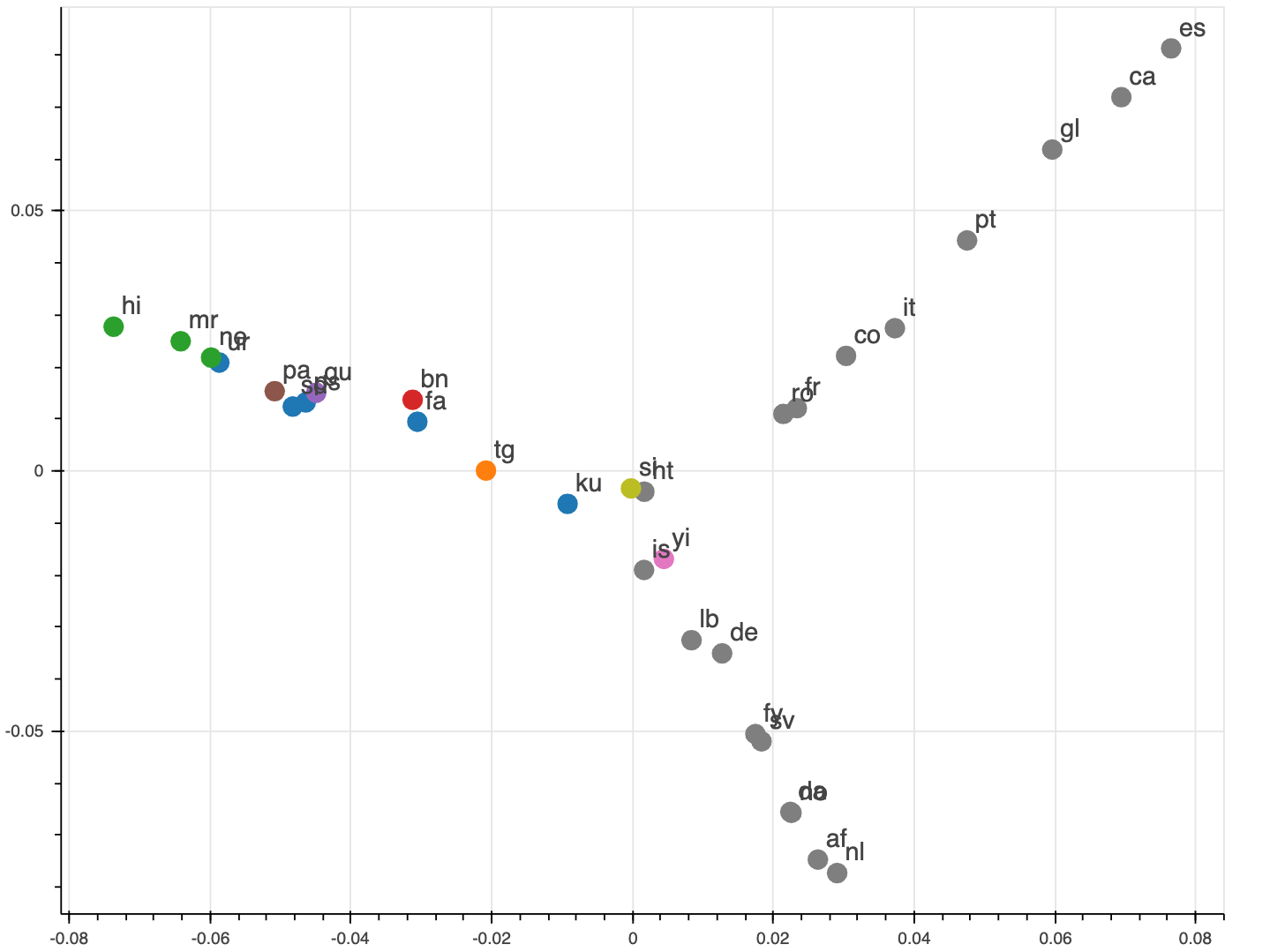}
    \captionsetup{width=.8\linewidth}
  \caption{Colored by script: Roman script in grey, Arabic script in blue, and a variety of other (mostly Indic) scripts.}
  \end{subfigure}
\caption{Visualization of the embedding layer for three branches of the Indo-European language family, coloring by different attributes to highlight clusters. They group most strongly by linguistic group, with weak connection by script.}
\label{fig:indoeuro-clusters}

\end{figure*}

In this section we further the analysis from subsection  \ref{section:linguistic-clusters}, with a different set of language families. In Figure \ref{fig:indo-iranian-dravidian-clusters}, we visualize the relationship between representations of the Iranian, Indo-Aryan, and Dravidian languages, and demonstrate that they cluster much more strongly by linguistic similarity than by script or dataset size. We furthermore demonstrate that within macro-clusters corresponding to languages of particular families, there exist micro-clusters corresponding to branches within those families.

This is a diverse set of mid-to-low resource languages, using a variety of scripts. The Dravidian languages (Kannada, Malayalam, Tamil, and Telugu) are from South India, and each use their own abugida-based writing system. They are agglutinative. The Indo-Aryan languages comprise the North-Indian languages, which are fusional languages written in Devanagari (Hindi, Marathi, Nepali), Arabic script (Sindhi, Urdu), and several language-specific abugidas (Bengali, Gujarati, Punjabi). The Iranian languages include Farsi (Dari), Kurdish (in this case, Kurmanji), and Pashto, written in Arabic script; and Tajik, written in Cyrillic. All languages in these three groups use SOV word order, and lie along swath of land stretching roughly from Sri Lanka to Kurdistan. In our datasets they are all low-resource languages, though Hindi is on the upper end.

The first most striking thing about Figure \ref{fig:indo-iranian-dravidian-clusters} is that \textbf{the linguistic group of each language appears to be a much stronger influence on the clustering tendency than the script, even at the level of the embeddings.} The Dravidian languages cluster nicely, even though none shares a subword with the other; as do the Indo-Aryan languages, which similarly are written in a variety of scripts, and the Iranian languages, which are written in two.

\begin{figure*}[ht!]
\begin{subfigure}{0.5\textwidth}
  \centering
  \includegraphics[width=.8\linewidth]{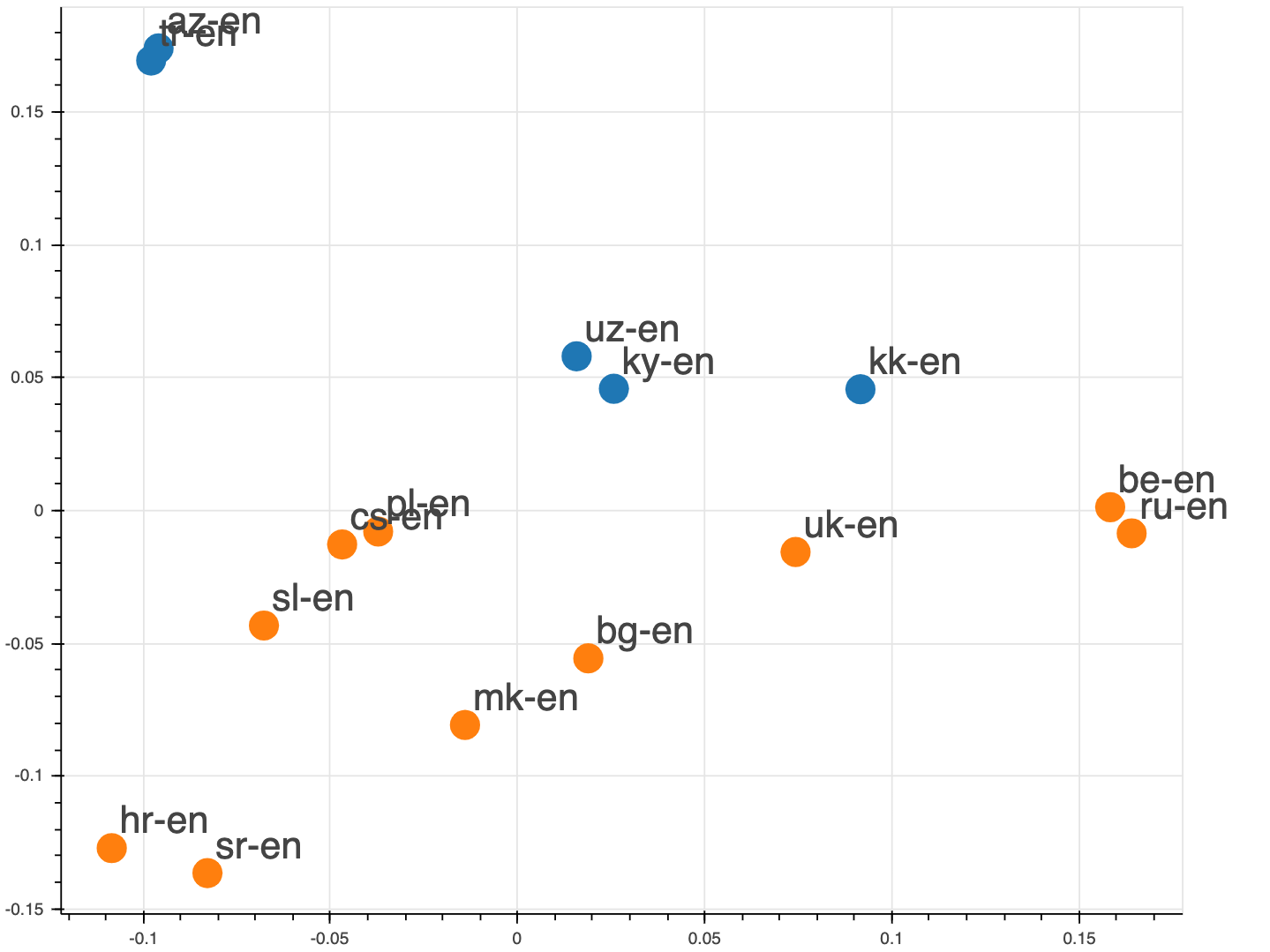}
  \caption{Embeddings, colored by sub-family}
\label{fig:slavoturkic-clusters:emb-script}
  \end{subfigure}
\begin{subfigure}{0.5\textwidth}
  \centering
  \includegraphics[width=.8\linewidth]{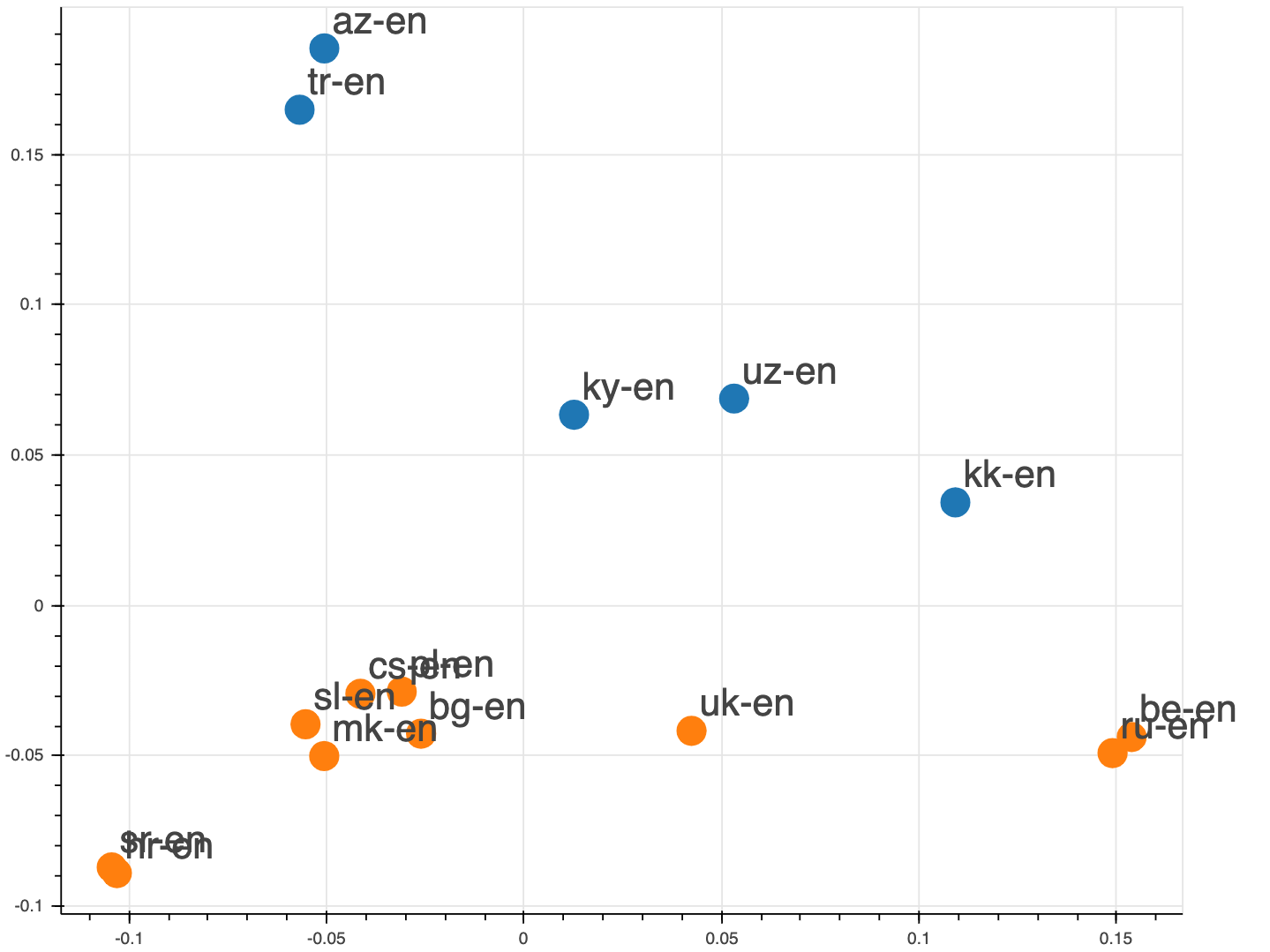}
  \caption{Encoder layer 5, colored by sub-family}
\label{fig:slavoturkic-clusters:enc5-script}
  \end{subfigure}
\begin{subfigure}{0.5\textwidth}
  \centering
  \includegraphics[width=.8\linewidth]{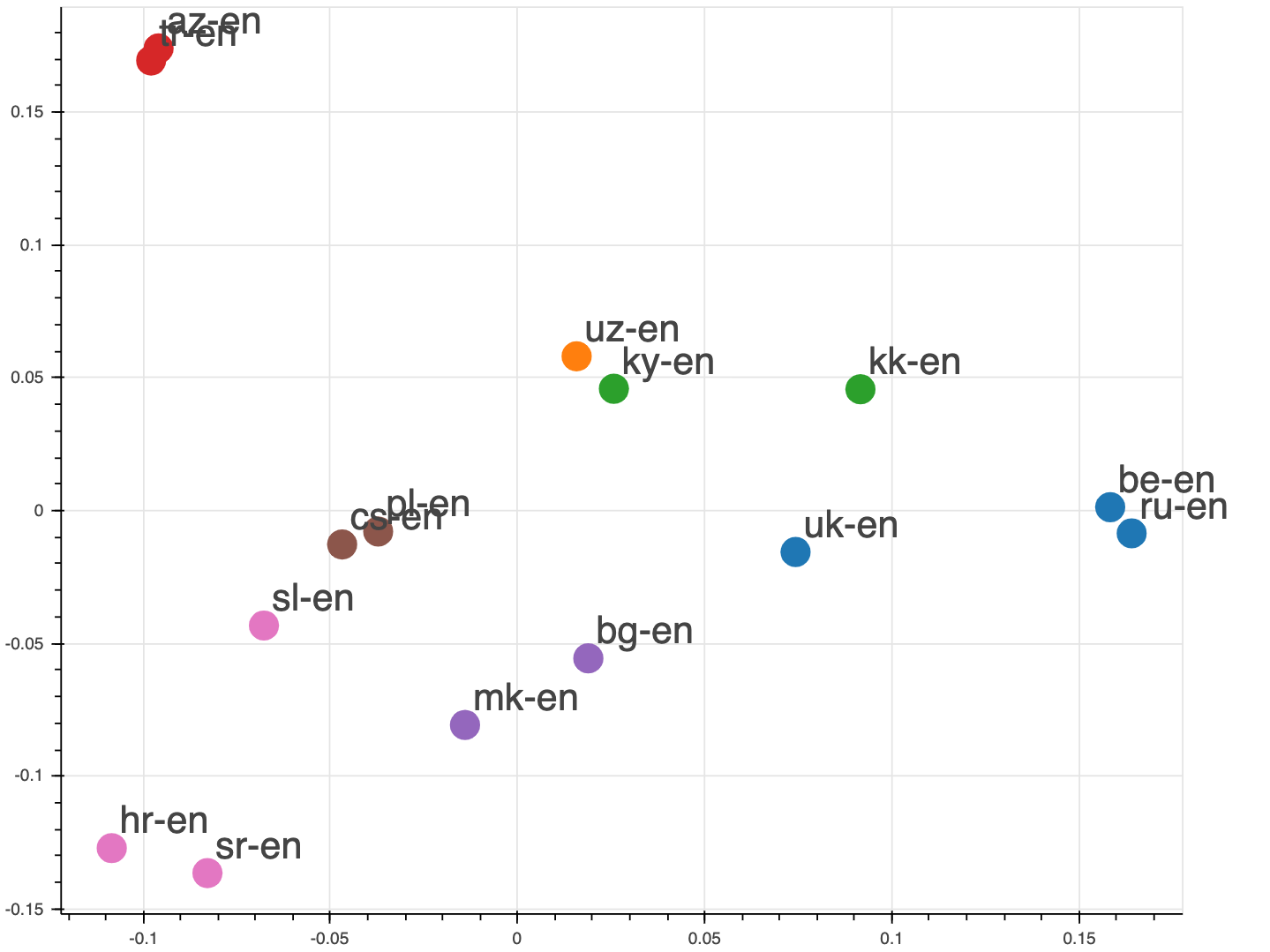}       
  \caption{Embeddings, colored by branch with sub-family}
\label{fig:slavoturkic-clusters:emb-fam}
  \end{subfigure}
\begin{subfigure}{0.5\textwidth}
  \centering
  \includegraphics[width=.8\linewidth]{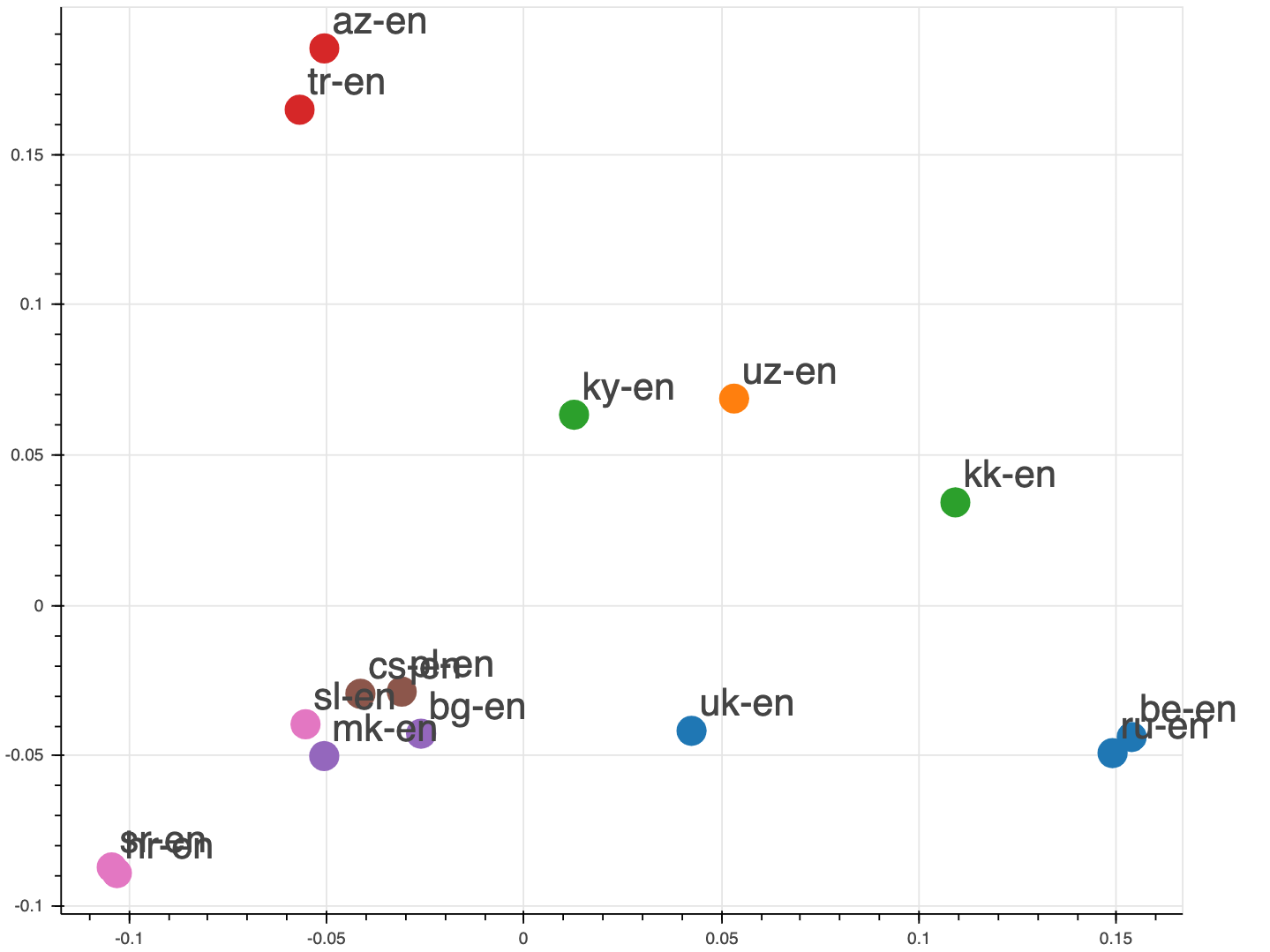}       
  \caption{Encoder layer 5, colored by branch within sub-family}
\label{fig:slavoturkic-clusters:enc5-fam}
  \end{subfigure}
\begin{subfigure}{0.5\textwidth}
  \centering
  \includegraphics[width=.8\linewidth]{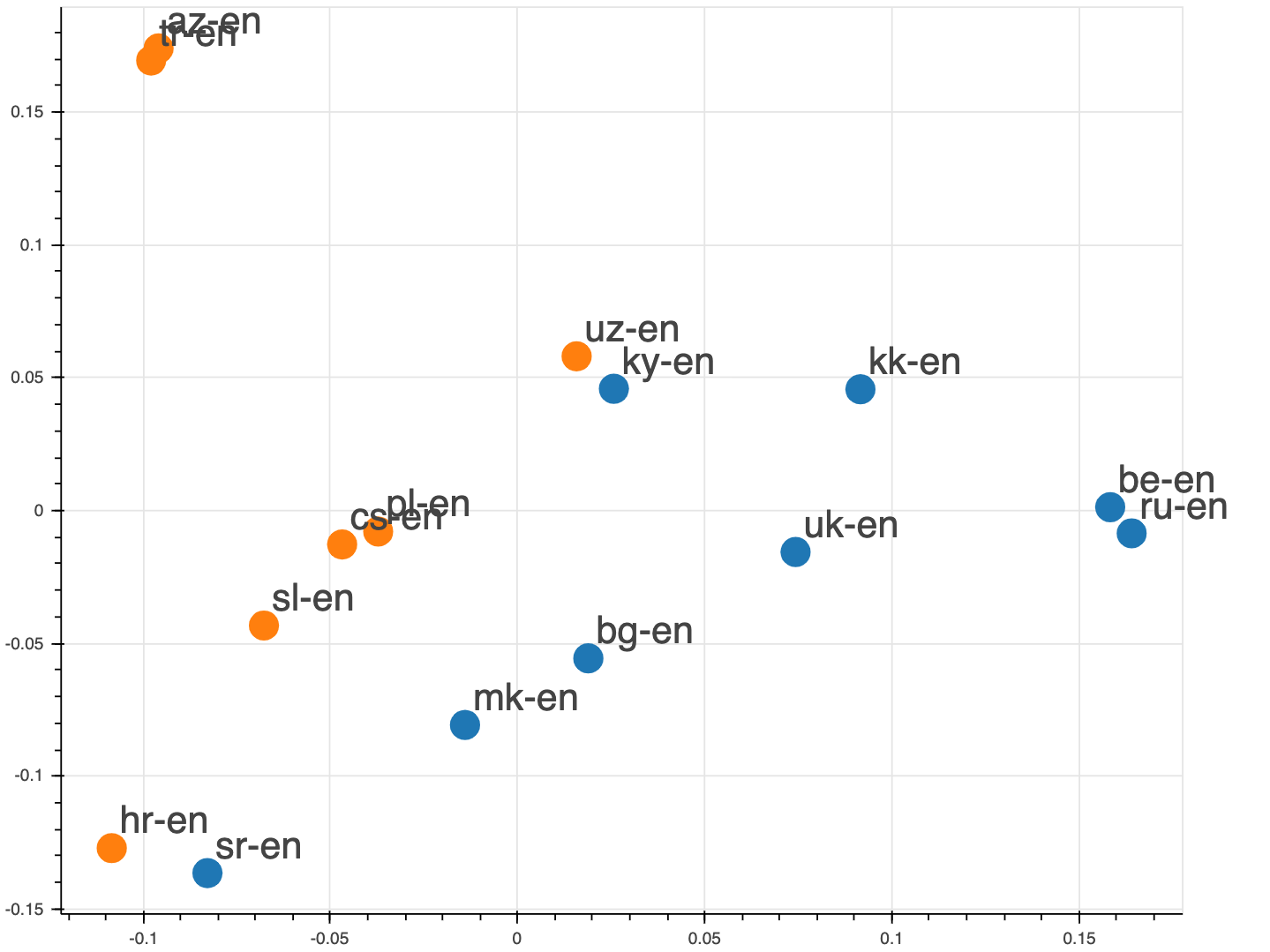}       
  \caption{Embeddings, colored by script (writing system)}
\label{fig:slavoturkic-clusters:emb-resource}
  \end{subfigure}
\begin{subfigure}{0.5\textwidth}
  \centering
  \includegraphics[width=.8\linewidth]{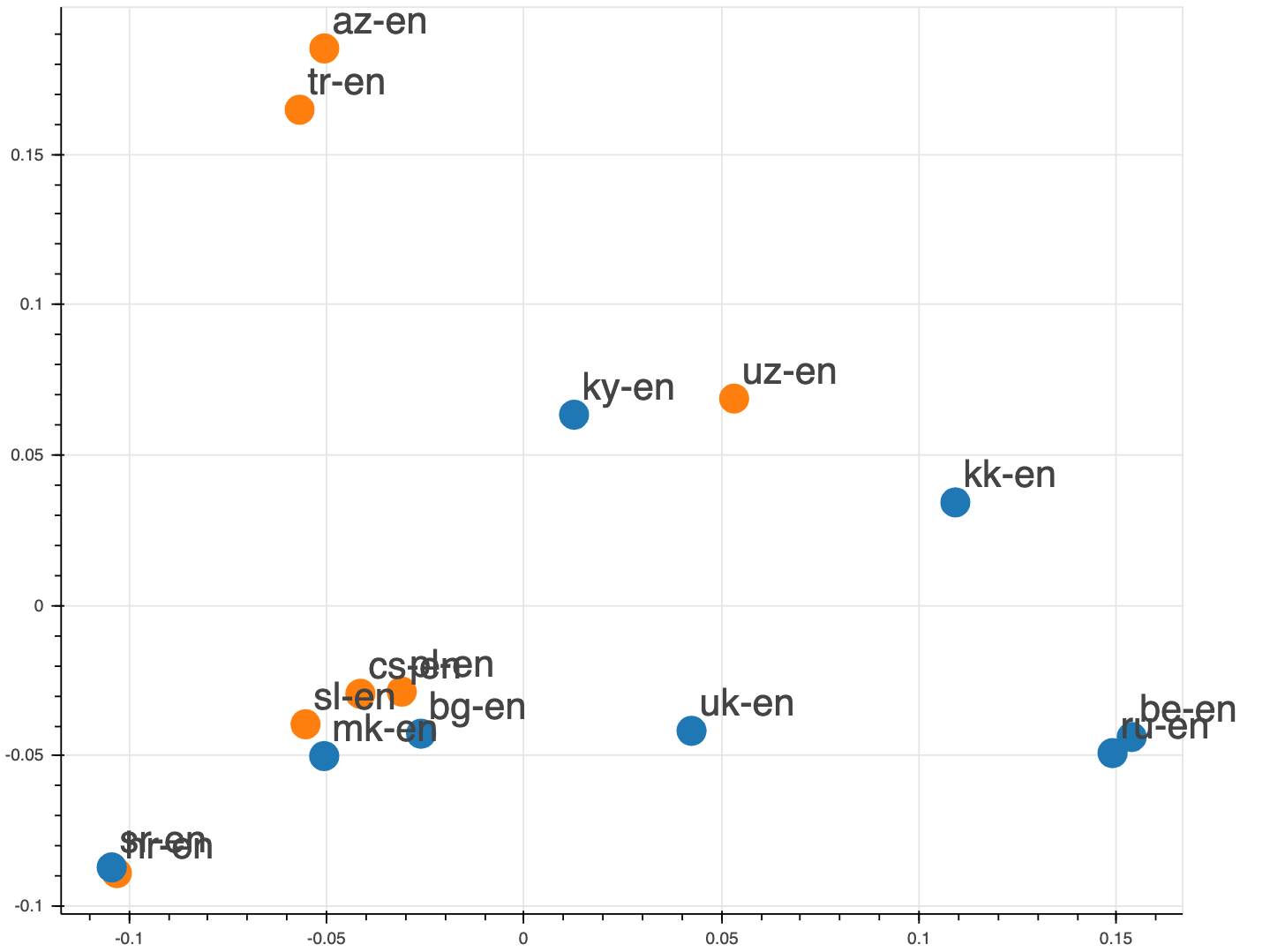}       
  \caption{Encoder layer 5, colored by script (writing system)}
\label{fig:slavoturkic-clusters:enc5-resource}
  \end{subfigure}
  
\caption{Visualization of the Slavic and Turkic languages, for the embeddings (left column) and the top layer of the encoder (right column), coloring by different attributes to highlight clusters. These are to-English direction.}
\label{fig:slavoturkic-clusters}
\end{figure*}

It is worth highlighting a few phenomena in this visualization. Firstly, the two dialects of Persian represented here, Tajik (tg) and Farsi (fa), are almost superimposed in the embedding visualization, though they are written in Cyrillic and Arabic scripts, respectively. Note also that in the embeddings, Hindi (hi) clusters closeliest with its fellow Western Indo-Aryan language Gujarati (gu), rather than the two other languages written in the Devanagari script, Marathi (mr) and Nepali (ne). Among the Dravidian languages, we see that Kannada (kn) and Telugu (te) form one pair, whereas Tamil (ta) and Malayalam (ml) form another pair, corresponding correctly with their linguistic similarity \cite{protodravidian}, even though none of them share a writing system with any of the others.

Although the language family is important to these clusters, it is important to note the apparent role that writing system also plays in these visualizations. While Urdu (ur) and Sindhi (sd) weakly cluster with the Indo-Aryan languages, they are as close in the visualization to the less related Iranian languages (which also use the Arabic script) than their linguistic nearest neighbors, Hindi (hi) \footnote{Usually considered to be a register of the same language \cite{hindiurdu1, hindiurdu3} } and Punjabi (pa) \footnote{In that Sindhi and Punjabi are the two representatives of the Northwestern Indo-Aryan languages in this plot.} , respectively.

\subsubsection*{Mid to high-resource, same-scripted languages}
Unsurprisingly, in the higher-resource case, and when most languages use a comparable script, the clusters are much cleaner. Figure \ref{fig:indoeuro-clusters} shows an example with the Romance languages and the the Germanic languages, along with the Indo-Aryan languages for comparison. They form three distinct clusters along what appear to be latent directions encoding language family. The apparent intersection off the three contains a few low-resource languages, and can best be conceptualized as an artifact of the visualization.

\subsection{Nearest neighbors based on representation similarity} \label{sub:nn}

In this section we look at how the nearest neighbors to languages change from their representations in the embeddings to their representations in the top layer of the encoder. Table \ref{nns} displays a few representative results.

The nearest neighbors of languages in high-resource language families, like Romance languages and Germanic languages, tend to produce quite accurate nearest neighbor lists that are stable across layers in the encoder. The example of Spanish in Table \ref{nns} demonstrates this, producing (at the encoder top) a remarkable list of the five linguistically closest languages to it in our dataset.

Lower-resource languages, however, tend to produce much noisier representations in the embeddings. The example of Yiddish is given in Table \ref{nns}. The nearest neighbors in the embedding space are mostly nonsensical, with the exception of German. By the top of the encoder, however, the neighbors are really quite reasonable, and remarkable -- given that Yiddish, a Germanic language \cite{yiddish}, is written in Hebrew script, whereas the remainder of the Germanic languages are written in Roman script. A similar example is Urdu, where the embeddings seem to be more influenced by less-related languages written in the same (Arabic) script, whereas by the top of the encoder, the neighbor list is a quite high-quality ranking of similar languages in entirely different scripts.

A last amusing example is Basque, a famous language isolate hiding out amidst the Indo-European languages in Europe. As expected from its status as an isolate, the nearest neighbors in the embedding space are a nonsensical mix of languages. However, by the top of the encoder the top four nearest neighbors are those languages geographically closest to Basque country (excepting French), probably reflecting lexical borrowing or areal influences on Basque.

\begin{table}[ht]
\small
\centering
\begin{tabular}{|l|c|l|l|}
\hline
Language & Rank & embeddings & encoder top \\
 \hline\hline
Yiddish & 1 & \textit{Lao} &  \textit{German} $\star\star$ \\ 
& 2 & \textit{German} $\star\star$ & \textit{Norwegian} $\star$ \\
& 3 & \textit{Thai} & \textit{Danish} $\star$ \\
& 4 & \textit{Hmong} & \textit{Portuguese} \\
& 5 & \textit{Korean} & \textit{Macedonian} \\
 \hline
Urdu & 1 & \textit{Punjabi} $\star$ & \textit{Hindi} $\star\star$\\
& 2 & Sindhi $\star$  & \textit{Punjabi} $\star$ \\
& 3 & Pashto & \textit{Bengali} $\star$ \\
& 4 & \textit{Hindi} $\star\star$ &  \textit{Gujarati} $\star$ \\
& 5 & \textit{Gujarati} $\star$ & \textit{Marathi} $\star$ \\
\hline
Basque & 1 & Indonesian & Portuguese \\
& 2 & Javanese & Spanish \\
& 3 & Portuguese & Galician \\
& 4 & Frisian & Italian  \\
& 5 & Norwegian & \textit{Bosnian} \\
\hline
Spanish & 1 & Catalan $\star$ & Catalan $\star$ \\
& 2 & Galician $\star$ & Portuguese $\star$ \\
& 3 & Portuguese $\star$ & Galician $\star$   \\
& 4 & Italian $\star$ & Italian  $\star$ \\
& 5 & Romanian $\star$ & French $\star$ \\
\hline 
\end{tabular}
\caption{Nearest Neighbors in SVCCA space for a given source language. Languages marked with a $\star$ are closely related; languages marked with $\star \star$ are the closest languages in our dataset. Italicized languages are written in a different script. We see that the nearest neighbors are more meaningful in the top of the encoder than in the embeddings, and that the embeddings are more influenced by script. \label{nns}}

\end{table}

In this example, the clusters remain about the same throughout the encoder, with the linguistic clusters becoming if anything a little tighter by the top layer of the encoder (Figure  \ref{fig:indo-iranian-dravidian-clusters:enc5-fam}, \ref{fig:indo-iranian-dravidian-clusters:enc5-script}). Sindhi and Urdu remain between the Indo-Aryan and Iranian languages. The one notable difference is that, whereas Sinhala (si) clusters with the Indo-Aryan languages in the embedding layer, it is firmly in the Dravidian cluster in the top of the encoder, with its nearest neighbor being Tamil. This may reflects the status of Sinhala as an Indo-Aryan language which has been lexically and grammatically influenced by sharing the island of Sri Lanka with Tamil over a thousand of years, to the extent that some earlier scholars erroneously believed the language to be Dravidian \cite{sinhalatamil}. Alternately it could reflect similar subject matter of text, related to e.g. local politics -- further analysis is required.

\subsection{Finetuning Experiments}\label{sub:ft}

In Table \ref{tab:lps} we list the language pairs with which we separately finetune our models and their resource sizes.  

\subsubsection*{Resource Size of Fine-tuning Language Pairs}\label{sub:ft-size}

\begin{table}[h]
    \centering
    \begin{tabular}{|p{4.1cm}|p{2.5cm}|}
\hline
Resource Size & Languages \\\hline
 Low ($10^5-10^7$ sentences)  & mr-en, km-en, uz-en, so-en, ky-en, ny-en, yo-en, ha-en, gd-en, ig-en  \\
High ($10^8-10^9$ sentences) & es-en, tr-en,pl-en, ko-en, ru-en, sr-en, uk-en, ca-en  \\
\hline

    \end{tabular}
    \caption{Language pairs we finetune our model on. For the purpose of our analysis, low resource languages are language pairs whose training set contained $10^5-10^7$ parallel sentences and high resource languages are languages pairs  whose training set contained $10^8-10^9$ parallel sentences.} 
    \label{tab:lps}
\end{table}

\subsubsection*{Sensitivity to Fine-tuning Increases Across Layers}\label{sub:ft-var}

\begin{figure*} [ht]
\begin{center}
\begin{subfigure}{0.45\textwidth}
  \centering
  \includegraphics[width=.9\linewidth]{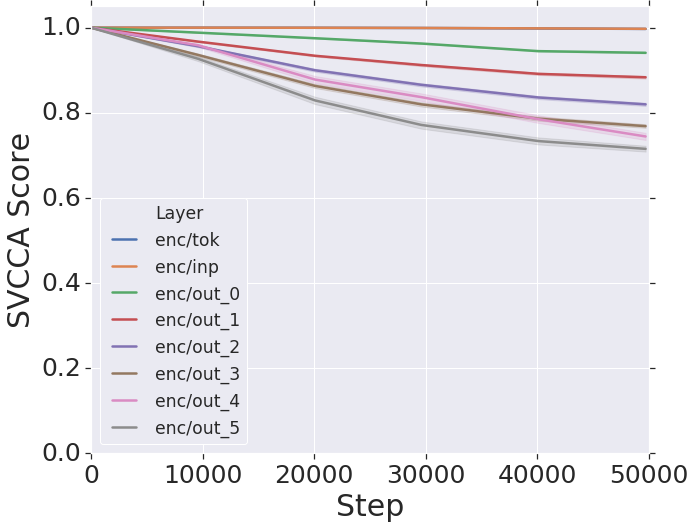}       
  \caption{Change in encoder on finetuning with ru-en.}
\label{fig:ruen-enc}
  \end{subfigure}
\begin{subfigure}{0.45\textwidth}
  \centering
  \includegraphics[width=.9\linewidth]{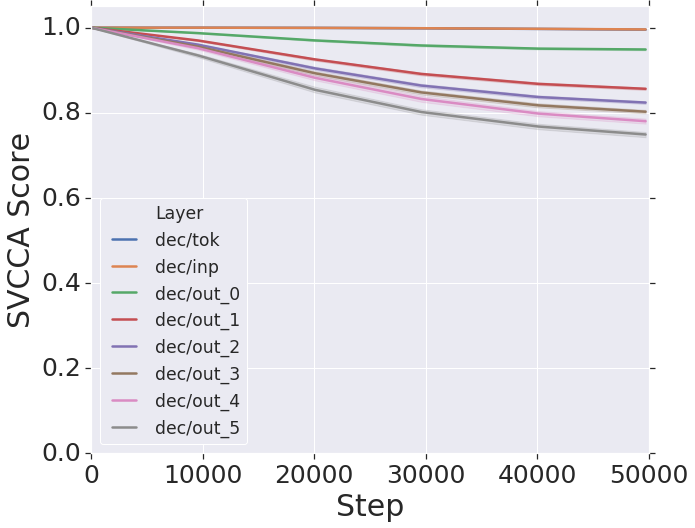}       
  \caption{Change in decoder on finetuning with ru-en.}
\label{fig:ruen-dec}
  \end{subfigure}
\begin{subfigure}{0.45\textwidth}
  \centering
  \includegraphics[width=.9\linewidth]{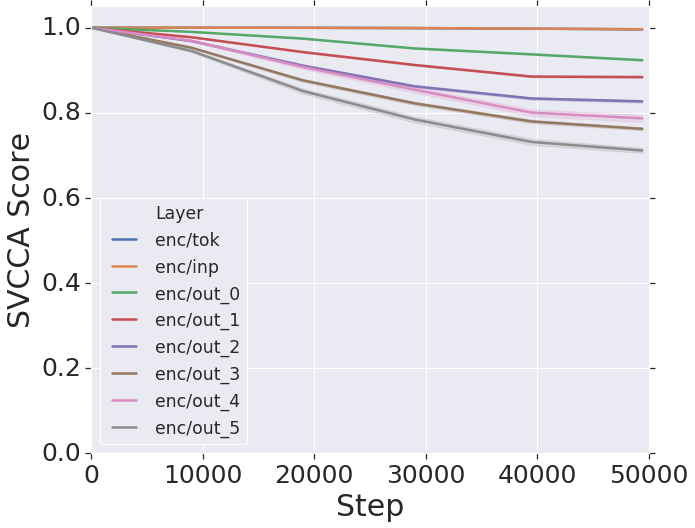}       
  \caption{Change in encoder on finetuning with ko-en.}
\label{fig:koen-enc}
  \end{subfigure}
\begin{subfigure}{0.45\textwidth}
  \centering
  \includegraphics[width=.9\linewidth]{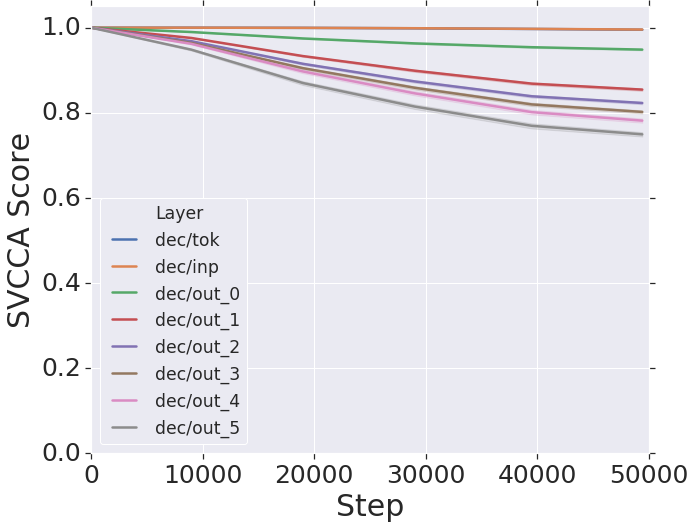}       
  \caption{Change in decoder on finetuning with ko-en.}
\label{fig:koen-dec}
  \end{subfigure}
\begin{subfigure}{0.45\textwidth}
  \centering
  \includegraphics[width=.9\linewidth]{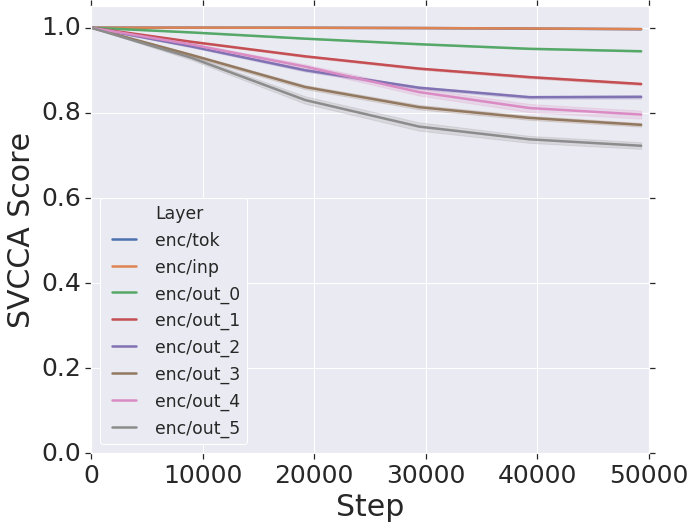}       
  \caption{Change in encoder on finetuning with uk-en.}
\label{fig:uken-enc}
  \end{subfigure}
\begin{subfigure}{0.45\textwidth}
  \centering
  \includegraphics[width=.9\linewidth]{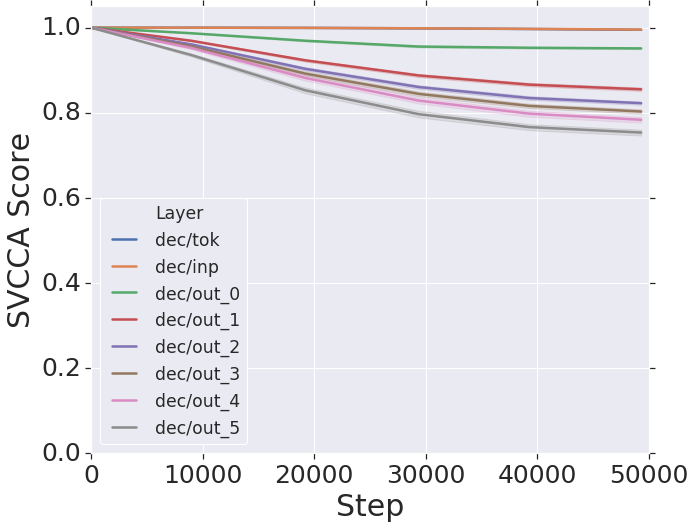}       
  \caption{Change in decoder on finetuning with uk-en.}
\label{fig:uken-dec}
  \end{subfigure}
\begin{subfigure}{0.45\textwidth}
  \centering
  \includegraphics[width=.9\linewidth]{pics/ruen_enc_ft.png}       
  \caption{Change in encoder on finetuning with km-en.}
\label{fig:kmen-enc}
  \end{subfigure}
\begin{subfigure}{0.45\textwidth}
  \centering
  \includegraphics[width=.9\linewidth]{pics/ruen_dec_ft.png}       
  \caption{Change in decoder on finetuning with km-en.}
\label{fig:kmen-dec}
  \end{subfigure}
\caption{Comparing average change in representation space over finetuning steps across layers for various language pairs.}
\label{fig:ft-ld}
\end{center}
\end{figure*}

In this subsection we plot the extent to which the representation space changes on average across language pairs (ie, decrease in SVCCA score) for different layers in Figure \ref{fig:ft-ld} on finetuning with these language pairs: ru-en (Russian), ko-en (Korean), uk-en (Ukrainian), km-en (Khmer). We see that the latter layers change the most across both the encoder and decoder.

\begin{table*}[ht]
\begin{tabular}{ll|ll|ll|ll}
\hline Language         & Id  & Language      & Id  & Language      & Id  & Language      & Id \\ \hline \hline
Afrikaans        & af & Galician         & gl & Latvian       & lv & Sindhi        & sd \\
Albanian         & sq & Georgian         & ka & Lithuanian    & lt & Sinhalese     & si \\
Amharic          & am & German           & de & Luxembouish   & lb & Slovak        & sk \\
Arabic           & ar & Greek            & el & Macedonian    & mk & Slovenian     & sl \\
Armenian         & hy & Gujarati         & gu & Malagasy      & mg & Somali        & so \\
Azerbaijani      & az & Haitian Creole    & ht & Malay         & ms & Spanish       & es \\
Basque           & eu & Hausa            & ha & Malayalam     & ml & Sundanese     & su \\
Belarusian       & be & Hawaiian        & haw & Maltese       & mt & Swahili       & sw \\
Bengali          & bn & Hebrew           & iw & Maori         & mi & Swedish       & sv \\
Bosnian          & bs & Hindi            & hi & Marathi       & mr & Tajik         & tg \\
Bulgarian        & bg & Hmong           & hmn & Mongolian     & mn & Tamil         & ta \\
Burmese          & my & Hungarian        & hu & Nepali        & ne & Telugu        & te \\
Catalan          & ca & Icelandic        & is & Norwegian     & no & Thai          & th \\
Cebuano         & ceb & Igbo             & ig & Nyanja        & ny & Turkish       & tr \\
Chinese          & zh & Indonesian       & id & Pashto        & ps & Ukrainian     & uk \\
Corsican         & co & Irish            & ga & Persian       & fa & Urdu          & ur \\
Croatian         & hr & Italian          & it & Polish        & pl & Uzbek         & uz \\
Czech            & cs & Japanese         & ja & Portuguese    & pt & Vietnamese    & vi \\
Danish           & da & Javanese         & jw & Punjabi       & pa & Welsh         & cy \\
Dutch            & nl & Kannada          & kn & Romanian      & ro & Xhosa         & xh \\
Esperanto        & eo & Kazakh           & kk & Russian       & ru & Yiddish       & yi \\
Estonian         & et & Khmer            & km & Samoan        & sm & Yoruba        & yo \\
Filipino/Tagalog  & tl & Korean           & ko & Scots Gaelic & gd   & Zulu          & zu \\
Finnish          & fi & Kurdish          & ku & Serbian       & sr & & \\
French           & fr & Kyrgyz           & ky & Sesotho       & st & & \\
Frisian          & fy & Lao              & lo & Shona         & sn & & \\ \hline
\end{tabular}
\caption{List of BCP-47 language codes used throughout this paper \cite{bcp47}.}.
\label{tab:langids}
\end{table*}

\end{document}